\documentclass[lettersize,journal]{IEEEtran}
\usepackage{booktabs}
\usepackage{cite}
\usepackage{amsmath,amssymb,amsfonts,amsthm}
\usepackage{algorithmic}
\usepackage{algorithm}
\usepackage{graphicx}
\usepackage{textcomp}
\usepackage{url}
\usepackage{subfigure}
\usepackage{enumitem}
\usepackage{array} 
\usepackage{pifont}  
\usepackage[table]{xcolor}
\usepackage{makecell}
\newcommand{\rev}[1]{\textcolor{black}{#1}}
\providecommand{\mycmark}{\ding{51}}
\providecommand{\myxmark}{\ding{55}}

\begin{document}
\bstctlcite{BSTcontrol}
\vspace{-0.2in}
\title{A General Deep Learning Framework for Wireless Resource Allocation under Discrete Constraints}
\author{Yikun Wang, Yang Li, Yik-Chung Wu, and Rui Zhang
\vspace{-0.2in}
\thanks{Y. Wang is with the Department of Electrical and Electronic Engineering, The University of Hong Kong, Hong Kong, and also with the School of Computing and Information Technology, Great Bay University, Dongguan 523000, China (email: ykwang@eee.hku.hk).}
\thanks{Y. Li is with the School of Computing and Information Technology, Great Bay University, Dongguan 523000, China (email: liyang@gbu.edu.cn). }
\thanks{Y.-C. Wu is with the Department of Electrical and Electronic Engineering, The University of Hong Kong, Hong Kong (email: ycwu@eee.hku.hk).}
\thanks{R. Zhang is with the Department of Electrical and Computer Engineering, National University of Singapore, Singapore (email: elezhang@nus.edu.sg).}}

\maketitle

\begin{abstract}
While deep learning (DL)-based methods have achieved remarkable success in continuous wireless resource allocation, efficient solutions for problems involving discrete variables remain challenging. This is primarily due to the zero-gradient issue in backpropagation, the difficulty of enforcing intricate constraints with discrete variables, and the inability in generating solutions with non-same-parameter-same-decision (non-SPSD) property. To address these challenges, this paper proposes a general DL framework by introducing the support set to represent the discrete variables. We model the elements of the support set as random variables and learn their joint probability distribution. By factorizing the joint probability as the product of conditional probabilities, each conditional probability is sequentially learned. This probabilistic modeling directly tackles all the aforementioned challenges of DL for handling discrete variables. By operating on probability distributions instead of hard binary decisions, the framework naturally avoids the zero-gradient issue. During the learning of the conditional probabilities, discrete constraints can be seamlessly enforced by masking out infeasible solutions. Moreover, with a dynamic context embedding that captures the evolving discrete solutions, the non-SPSD property is inherently provided by the proposed framework. We apply the proposed framework to two representative mixed-discrete wireless resource allocation problems: (a) joint user association and beamforming in cell-free systems, and (b) joint antenna positioning and beamforming in movable antenna-aided systems. Simulation results demonstrate that the proposed DL framework consistently outperforms existing baselines in terms of both system performance and computational efficiency.
\end{abstract}

\vspace{-0.3in}
\section{Introduction}
Mixed-discrete optimization problems are pervasive in wireless resource allocation, particularly in emerging paradigms such as cell-free (CF) systems \cite{9309348}, holographic multiple-input multiple-output (MIMO) systems\cite{9136592}, and fluid antenna (FA) or movable antenna (MA)-aided systems\cite{9264694,10354003}. In these applications, optimization involves not only continuous variables such as beamforming vectors and transmit powers, but also discrete decisions including user scheduling, antenna selection, or antenna positioning. Mathematically, these problems are commonly NP-hard, making it difficult to find the optimal solution~\cite{survey_co}. 

A common heuristic approach is to solve the problem in two steps, i.e., the discrete variables are first determined using rule-based or greedy methods~\cite{Heuristic}, and then the continuous variables are optimized under the fixed discrete variable configuration. However, such two-stage strategies suffer from severe performance loss due to the absence of joint optimization. To address this issue, several joint optimization algorithms have been proposed \cite{SWMMSE, FPII, C_FP_MA,11048972, YangYizhen2022, YangYizhen2024}. Many of these methods rely on continuous relaxation of discrete variables to enable gradient-based optimization \cite{SWMMSE,C_FP_MA,11048972}. However, such relaxation inevitably introduces a performance loss due to the relaxation gap and the need for post-hoc projection onto the discrete feasible set. Moreover, these approaches commonly require iterative updates \cite{SWMMSE, FPII, C_FP_MA,11048972, YangYizhen2022, YangYizhen2024}, which incur prohibitive computational latency \cite{engnn,9505260} and render them unsuitable for real-time deployment.

Recently, learning-to-optimize has emerged as a promising alternative, leveraging neural networks (NNs) to learn end-to-end mappings from environmental inputs to continuous resource allocation decisions\rev{~\cite{sun2026comprehensive}}. Representative architectures such as graph neural networks (GNNs) \cite{9505260,MDGNN,10970427} and transformers \cite{HPEtransformer,9941253,10707363} have demonstrated the ability to achieve near-optimal performance with remarkably low inference latency. However, directly extending these approaches to mixed-discrete optimization problems remains non-trivial, primarily due to the following three challenges.

The first challenge is due to the conflict between discrete outputs and gradient-based training. Standard NNs rely on backpropagation, which fails when outputs are discrete because the resulting gradients are zero almost everywhere \cite{VQVAE}. Several methods have been proposed to address this challenge.
One popular technique is the straight-through estimator (STE) \cite{PilotGNN, park2025self}, which bypasses the non-differentiability by copying gradients from the discrete outputs back to the continuous representation. However, such approach may provide misleading information about the true optimization landscape, causing performance degradation, especially when discrete variables are highly structured or coupled \cite{park2025self,Gsoftmax}. Alternative approaches approximate the discrete outputs using differentiable surrogates such as softmax or Gumbel-softmax \cite{Gsoftmax, LSJWideband, 11021397}. While these methods enable gradient flow, they introduce approximation errors that deviate from the true objective. Although a temperature parameter governs how well the relaxed output approximates a true discrete decision, it requires careful specification of the temperature schedule to balance the model learnability and approximation error.

The second challenge lies in difficulty in enforcing the constraints involving discrete variables in \rev{deep learning (DL)}. For instance, in MA-aided systems, to avoid mutual coupling between antennas, the distance between any pair of MAs must exceed a given threshold~\cite{10354003}. This yields a highly non-convex and combinatorial feasible set that is challenging for NNs to deal with. While penalty-based methods~\cite{Penalty} can incorporate such constraints into the loss function, they require meticulous tuning of penalty coefficients. Moreover, the penalty-based methods cannot guarantee strict feasibility, which is a critical limitation in applications with stringent reliability requirements.

The third challenge stems from the struggle of typical \rev{DL} in achieving non-same-parameter-same-decision (non-SPSD) property on the discrete variable solutions \cite{chen2025gnns, LSJWideband}. In wireless resource allocation, the problem is deemed non-SPSD when identical or nearly identical system parameters are assigned different solutions. For example, consider a multi-user scheduling problem, where two users experience nearly identical channel conditions. However, the optimal scheduler may activate only one of them due to the strong interference between them. This inherent asymmetry in the discrete solution cannot be captured by current DL approaches \cite{9505260,MDGNN,10970427,PilotGNN,11021397,HPEtransformer,9941253,10707363}, resulting in severe performance degradation \cite{chen2025gnns, LSJWideband}.

\rev{Although autoregressive pointer-based architectures have been designed for combinatorial optimization \cite{bello2016neural,kool2018attention}, directly applying these methods to wireless resource allocation is infeasible. Unlike conventional tasks such as the traveling salesman problem, wireless resource allocation exhibits distinct features, including mixed-variable types, highly coupled discrete constraints, unknown output cardinality, and non-SPSD properties. Consequently, substantial domain-specific innovations are required in wireless systems.}

To address the aforementioned challenges, we propose a general DL framework for a broad class of mixed-discrete wireless resource allocation problems. The proposed approach leverages a reformulation that expresses discrete decisions using a support set to enhance learnability. Building on this, we design a discrete variable learning network (DVLN) to learn the probability distribution over discrete solutions given system parameters. Crucially, to enforce the constraints with discrete variables, the DVLN adopts an encoder-decoder architecture with sequential decoding process: rather than predicting all discrete variables in parallel, it generates the solutions of different variables sequentially, dynamically masking infeasible candidates at each step to guarantee feasibility. Notably, the solution is generated by an attention mechanism based on a \emph{context embedding}, which takes into account the current discrete variable solution and evolves dynamically with each step, thereby enabling our DL framework to effectively capture the non-SPSD property. The decoder can also learn to adaptively terminates the decoding process, by predicting a learnable end-token. The DVLN is jointly trained with a continuous variable learning network (CVLN), which is specialized for outputting continuous resource allocation solutions. The two networks are optimized end-to-end in an unsupervised manner by directly maximizing the underlying system performance metric, without requiring a precomputed dataset of optimal solutions.

The main contributions of this paper are summarized as follows: 
\begin{itemize}[left=0pt]
\item A general formulation for mixed-discrete resource allocation in wireless systems is proposed, where discrete variables are represented using the support set. 
\item \rev{Inspired by the autoregressive pointer-based architectures~\cite{bello2016neural,kool2018attention}, we propose a general DL framework for mixed-discrete wireless resource allocation. Our framework comprises a DVLN and a CVLN. Specifically, the DVLN predicts the support set probabilities via sequential generation, wherein a constraint-aware masking strategy is incorporated for strict constraint enforcement, and a dynamic context embedding is utilized to handle non-SPSD conditions. A learnable end-token is designed to adaptively terminate the decoding process.} With the CVLN being flexible to tailor various continuous resource variables, the DVLN and CVLN are jointly trained in an unsupervised manner. 
\item The proposed DL framework is illustrated with two typical wireless examples: (a) joint user equipment (UE)–access point (AP) association and beamforming in CF systems; and (b) joint antenna positioning and beamforming in MA-aided systems. In both cases,
simulations show that the proposed DL framework achieves superior performance and strict constraint satisfaction with significantly lower inference latency than existing DL-based and model-based approaches.
\end{itemize}

The remaining paper is organized as follows. Section \uppercase\expandafter{\romannumeral2} introduces a general problem formulation and two application examples. Section \uppercase\expandafter{\romannumeral3} describes the general DL framework, while Section \uppercase\expandafter{\romannumeral4} introduces the training algorithm. In Sections \uppercase\expandafter{\romannumeral5} and \uppercase\expandafter{\romannumeral6}, the DL framework is demonstrated with two examples, with performance evaluated through simulations. Finally, the conclusion is drawn in Section \uppercase\expandafter{\romannumeral7}.

\textit{Notations:} We use $a$, $\bf{a}$, $\bf{A}$, and $\mathcal{A}$ to denote scalar, vector, matrix, and set, respectively; $\mathbf{A}^{T}$ and $\mathbf{A}^{H}$ to denote transpose and Hermitian transpose; $\cup$ and $\cap$ to denote the set union and intersection; $|\cdot|$ and $\|\cdot\|$ to denote the modulus and $\ell_2$ norm;  $\mathcal{CN}(\cdot,\cdot)$ to denote complex Gaussian distribution.

\section{General Problem Formulation and Application Examples}
\subsection{Problem Formulation}
Many wireless resource allocation problems appear as a mixed-discrete optimization problem in the following form:

\begin{subequations}
\vspace{-0.15in}
\label{problem1}
\begin{align}
 \text{(P1):} &\max _{\mathbf{b}\in\{0,1\}^{N_{\text{b}}},\mathbf{w}\in\mathbb{C}^{N_{\text{w}}}}& &U(\mathbf{b},\mathbf{w};{\mathbf{h}})&
\label{bin_obj} \\
&~~~~~~~\text{s.t.}& &f_{m}(\mathbf{b};\mathbf{h})\leq0,~\forall m=1,2,\dots, \Omega,& \label{binb}\\ 
&& & g_m(\mathbf{w};\mathbf{h})\leq0,~\forall m=1,2,\dots, \Psi,&\label{binc}
\vspace{-0.1in}
\end{align}
\end{subequations}
where $\mathbf{b}\in \{0,1\}^{N_{\text{b}}}$ and $\mathbf{w}\in \mathbb{C}^{N_{\text{w}}}$ denote the discrete and continuous variables, respectively, and $\mathbf{h}\in \mathbb{C}^{N_{\text{h}}}$ represents the known system parameters. Although the discrete variable $\mathbf{b}$ is written as a binary vector with each entry $b_n\in\{0,1\}$, this representation can handle vectors taking any finite discrete values  via binary encoding schemes like one-hot encoding (e.g., see \textbf{Example 2} in Subsection~C). 

Beside the discrete variable $\mathbf{b}$, the general problem formulation (P1) also includes continuous variable $\mathbf{w}$, thus (P1) is a ``mixed'' optimization. For instance, $\mathbf{b}$ may represent the indicator of user scheduling or antenna selection, $\mathbf{w}$ may denote the beamforming or power allocation, and $\mathbf{h}$ represents the channel state information or user locations. The objective of (P1) is to maximize a utility function $U(\cdot,\cdot;\cdot)$, which may depend on all the variables and system parameters. Moreover, $f_m(\cdot;\cdot)$ and $g_m(\cdot;\cdot)$ impose possible restrictions on $\bf{b}$ and $\bf{w}$, respectively\footnote{For constraints involving coupled discrete and continuous variables, our proposed DL framework can be integrated with a previously proposed penalty-dual learning framework \cite{PDL}. However, this extension is beyond the scope of the present paper.}. This formulation (P1) covers applications in many previous works as special cases\cite{9309348,SWMMSE,FPII,C_FP_MA,11048972,PilotGNN,LSJWideband,11021397}. Specific examples are given in Subsection~C.

\vspace{-0.1in}
\subsection{Equivalent Problem Transformation}
In this subsection, we transform (P1) into an equivalent form by leveraging the concept of support set, which enables the NN to better learn the desired solution mapping. Specifically, the binary variable $\mathbf{b}$ can be fully captured by its support set $\mathcal{A}$, defined as the set of indices corresponding to the non-zeros entries in $\mathbf{b}$. Instead of optimizing over the full $N_{\text{b}}$-dimensional vector $\mathbf{b}$, the problem becomes identifying the support set $\mathcal{A}$. The reformulation is formally given by the following property.

\emph{Property 1 (Reformulation of (P1)):} Any mixed-discrete optimization problem in the form of (P1) is equivalent to
\begin{subequations}
\label{problem2}
\begin{align}
 \text{(P2):}&\max _{\mathcal{A}\in{2}^{\mathcal{N}_{\text{b}}},\mathbf{w}\in\mathbb{C}^{N_{\text{w}}}}&&{U}(\mathbf{b},\mathbf{w};\mathbf{h})
\label{disa} \\
&~~~~~~~\text{s.t.}& &\text{(\ref{binb}), (\ref{binc}),} \nonumber\\
&& &b_{n}=\left\{\begin{array}{ll}
1,\quad \forall n \in \mathcal{A}, \\
0 ,\quad \forall n \notin \mathcal{A},
\label{disd}\end{array}\right. 
\end{align}
\end{subequations}
where $\mathcal{N_{\text{b}}}\triangleq\{1,2,\dots,N_{\text{b}}\}$, and   $2^{\mathcal{N_{\text{b}}}}$ denotes the power set of $\mathcal{N}_{\text{b}}$, i.e., the collection of its all $2^{{N}_{\text{b}}}$ possible subsets. The discrete variable $\bf{b}$ is fully determined by $\cal{A}$ through the relationship in (\ref{disd}). For example, if $\mathcal{A}=\emptyset$, we have ${\mathbf{b}}={\mathbf{0}}$; while if $\mathcal{A}=\mathcal{N}_\text{b}$, we have ${\mathbf{b}}={\mathbf{1}}$.

In order to avoid repeatedly solving (P2) in a case-by-case manner for different $\mathbf{h}$, we strive to design a mapping function given by
\begin{equation}
\label{totalDNN}
\{\mathcal{A},\mathbf{w}\}= \mathcal{F}(\mathbf{h}),
\end{equation}
where $\mathcal{F}(\cdot):\mathbb{C}^{N_{\text{h}}}\to 2^{\mathcal{N}_{\text{b}}}\times \mathbb{C}^{N_{\text{w}}}$ is represented by a NN, such that the average objective function~\eqref{disa} is maximized. The corresponding DL-based optimization problem  can be written as
\begin{subequations}
\vspace{-0.15in}
\label{jointP}
\begin{align}
\text{(P3):}~ &\max_{\mathcal{F}(\cdot)}& &\mathbb{E}_{\mathbf{h}}[{U}(\mathbf{b},\mathbf{w};\mathbf{h})] \tag{4}
\label{obj_of_SR} \\
& ~\text{s.t.}& &\text{(\ref{binb})},~\text{(\ref{binc})},~\text{(\ref{disd})},~\text{(\ref{totalDNN})}, &&\nonumber
\end{align}
\end{subequations}
where the expectation is taken over the probability distribution of $\mathbf{h}$.

\vspace{-0.1in}
\subsection{Examples in Wireless Communications}
Here, we present two typical examples of mixed-discrete optimization problems in wireless communications and demonstrate how they can be expressed in the forms of (P2) and (P3). 

\begin{figure}[t]
\centering
\subfigure[Joint UE-AP association and beamforming in CF systems.]{\includegraphics[width=0.24\textwidth]{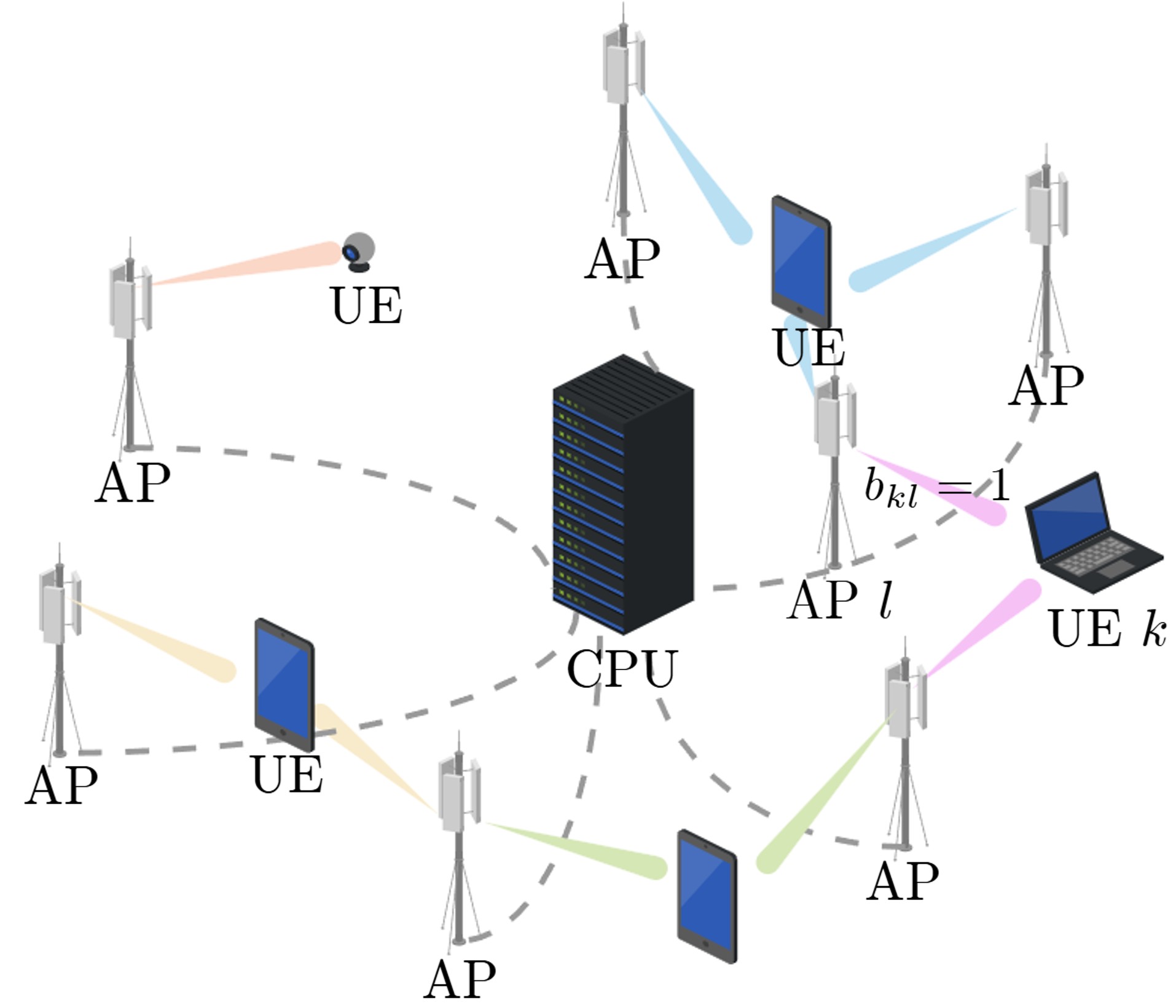}%
\label{CF_system}}
\subfigure[Joint antenna positioning and beamforming for MA-aided systems.]{\includegraphics[width=0.24\textwidth]{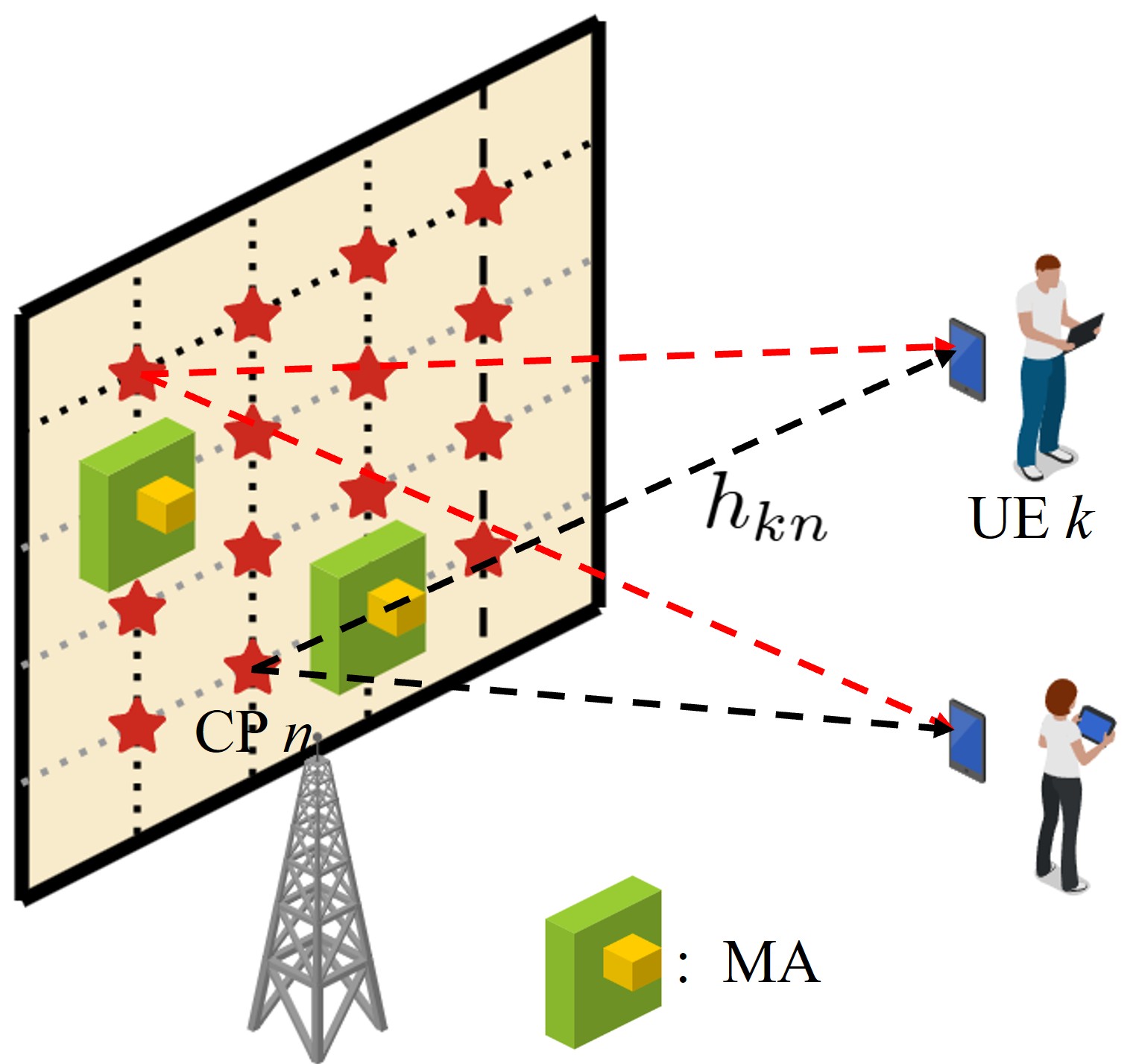}%
\label{MA_system}}
\caption{Illustrations of two typical examples of mixed-discrete wireless resource allocation problems.}
\label{Wholesystem}
\vspace{-0.2in}
\end{figure}

\textbf{Example 1 }\textit{(Joint UE-AP Association and Beamforming in CF Systems)}: Consider a downlink CF system comprising $L$ APs and $K$ UEs. Each UE is equipped with a single antenna and each AP is with $M$ antennas. All APs are connected to a central process unit (CPU) for joint signal processing (see Fig.~\ref{Wholesystem}(a)). Define the association variable as ${b_{kl}}\in\{0,1\}$, where ${b_{kl}}=1$ denotes that the $k$-th UE is served by the $l$-th AP, and ${b_{kl}}=0$ otherwise. Additionally, due to the limitation of fronthaul capacity, the following constraints are imposed on the association variables \cite{9443536,9309348}. First, each AP can at most serve $K_{\text{max}}\ll K$ UEs, which is mathematically described by
\begin{equation}
\vspace{-0.1in}
\label{equ6}
\sum\nolimits_{k=1}^{K}{b_{kl}} \leq K_{\text{max}}, ~~~\forall l\in\mathcal{L}\triangleq\{1,2,\dots,L\}. 
\end{equation}
Second, each UE is served by at most $L_{\text{max}}\ll L$ APs, which is given by
\begin{equation}
\vspace{-0.1in}
\label{equ7}
\sum\nolimits_{l=1}^{L}{b_{kl}} \leq L_{\text{max}}, ~~~\forall k\in\mathcal{K}\triangleq\{1,2,\dots,K\}.
\end{equation}

We adopt a block-fading channel model in which the channels are assumed to be constant within a coherence period. We define the channel between the $l$-th AP and the $k$-th UE as $\mathbf{h}_{kl}\in \mathbb{C}^{M}$. Consequently, the received signal at the $k$-th UE can be expressed as
\begin{equation}
\label{equ8}
y_k = \sum_{l=1}^{L}{b_{kl}}\mathbf{h}_{kl}^{H}\mathbf{w}_{kl}s_k+\sum_{k=1,k'\ne k}^{K}\sum_{l=1}^{L}b_{k'l}\mathbf{h}_{kl}^{H}\mathbf{w}_{k'l}s_{k'}+n_k,
\end{equation}
where $s_k \in \mathbb{C}$ denotes the desired symbol of the $k$-th UE that satisfies $\mathbb{E}[|s_k|
^2]=1$, $\mathbf{w}_{kl}\in \mathbb{C}^{M}$ is the beamforming vector of the $k$-th UE at the $l$-th AP, and $n_k\sim \mathcal{CN}(0,\sigma_k^2)$ is the additive white Gaussian noise (AWGN) with $\sigma_k^2$ being the noise power. From \eqref{equ8}, the achievable sum rate of all $K$ users can be expressed as
\begin{equation}
\vspace{-0.1in}
\begin{aligned}
\label{equ9}
& R^{\text{CF}}(\mathbf{b},\mathbf{w};\mathbf{h}) = \\
&\sum_{k=1}^{K}\log_2\left(1+\frac{|\sum_{l=1}^{L}{b_{kl}}\mathbf{h}_{kl}^{H}\mathbf{w}_{kl}|^2}{\sum_{k=1,k'\ne k}^{K}|\sum _{l=1}^{L}b_{k'l}\mathbf{h}_{kl}^{H}\mathbf{w}_{k'l}|^2+\sigma^2_k}\right),
\end{aligned}
\end{equation}
where $\mathbf{b}=[{b_{11}},{b_{12}},\dots,{b_{1L}},{b_{21}},\dots,{b_{KL}}]^{T}\in \{0,1\}^{KL}$, $\mathbf{w}=[\mathbf{w}_{11}^{T},\mathbf{w}_{12}^{T},\dots,\mathbf{w}_{1L}^{T},\mathbf{w}_{21}^{T},\dots,\mathbf{w}_{KL}^{T}]^T\in\mathbb{C}^{KLM}$, and $\mathbf{h}=[\mathbf{h}_{11}^{T},\mathbf{h}_{12}^{T},\dots,\mathbf{h}_{1L}^{T},\mathbf{h}_{21}^{T},\dots,\mathbf{h}_{KL}^{T}]^T\in \mathbb{C}^{KLM}$. Consequently, the joint UE-AP association and beamforming problem can be formulated as

\begin{subequations}
\label{optim}
\begin{align}
\text{(P4):}&\max _{\mathbf{b}\in\{0,1\}^{KL},\mathbf{w}\in \mathbb{C}^{KLM}}& &R^{\text{CF}}(\mathbf{b},\mathbf{w};\mathbf{h})& &&\\
&~~~~~~~~~\text{s.t.}&&\text{(\ref{equ6})},~\text{(\ref{equ7})},&\nonumber\\
&& &\sum_{k=1}^{K} \|\mathbf{w}_{kl}\|^{2} \leq P_{\text{max}}, \forall l \in \mathcal{L},
\label{ccon}\end{align}
\end{subequations}
where $P_{\text{max}}$ denotes the maximum transmit power of each AP. The problem (P4) shares the same form with (P1), with $N_\text{b}=KL$, $N_\text{w}=KLM$, ${N_{\text{h}}}=KLM$, $\Omega=K+L$, and $\Psi=L$. Using \textit{Property 1}, problem (P4) can be rewritten as
\begin{subequations}
\vspace{-0.1in}
\label{problem5}
\begin{align}
\text{(P5):}&\max _{\mathcal{A}\in 2^{\mathcal{K}\times\mathcal{L}},\mathbf{w}\in\mathbb{C}^{KLM}}& &{R^\text{CF}}(\mathbf{b},\mathbf{w};{\bf{h}})&
\label{cdisa} \\
&~~~~~~~~\text{s.t.}& &\text{(\ref{equ6})},\text{ (\ref{equ7})},\text{ (\ref{ccon}),}& \nonumber\\
&& &{b_{kl}}=\left\{\begin{array}{ll}
1,~~\forall (k,l)\in \mathcal{A}, \\
0,~~\forall (k,l)\notin \mathcal{A},
\label{cdisd}\end{array}\right. 
\end{align}
\end{subequations}
where $\mathcal{K}\times\mathcal{L}$ denotes the Cartesian product of set $\mathcal{K}$ and $\mathcal{L}$, i.e., $\mathcal{K}\times\mathcal{L}=\{(k,l)|k\in \mathcal{K} ,~l \in \mathcal{L}\}$. 

Let $\mathcal{F}^{\text{CF}}(\cdot):\mathbb{C}^{KLM}\to 2^{\mathcal{\mathcal{K}\times\mathcal{L}}} \times \mathbb{C}^{KLM}$ represent the to-be-designed joint UE-AP association and beamforming mapping, the corresponding DL problem is expressed in the form of (P3) as
\begin{subequations}
\vspace{-0.1in}
\begin{align}
\text{(P6):}&~\max _{\mathcal{F}^{\text{CF}}(\cdot)}& &\mathbb{E}_{\bf{h}}[R^\text{CF}(\mathbf{b},\mathbf{w};{\bf{h}})]&\\
&~~\text{s.t.}& &\text{(\ref{equ6}), (\ref{equ7}), (\ref{ccon}), (\ref{cdisd}),}& \nonumber \\
&& & \{\mathcal{A},\mathbf{w}\}=\mathcal{F}^{\text{CF}}(\mathbf{h}). \label{CFLEARN}
\end{align}
\end{subequations}

\textbf{Example 2 }\textit{(Joint Antenna Positioning and Beamforming  for MA-aided Systems)}: 
Consider an MA-aided downlink system with one base station (BS) serving $K$ UEs. The BS is equipped with $M$ MAs and each UE is equipped with a single fixed-position antenna. The positions of the $M$ MAs can be adjusted simultaneously within a predefined two-dimensional rectangular area, which is divided into $N\gg M$ candidate positions (CPs) for placing the $M$ MAs (see Fig.~\ref{Wholesystem}(b)), with the coordinate of the $n$-th CP denoted by $\mathbf{p}_n = [x_n, y_n]^T$, where $x_n$ and $y_n$ represent the coordinates along the x-axis and y-axis, respectively.

Define a binary positioning vector as $\mathbf{b}\triangleq [b_1,b_2,\dots,b_{N}]^T \in \{0,1\}^{N}$, where $b_n=1$ indicates an MA is placed at the $n$-th CP, and $b_n=0$ otherwise. The vector $\mathbf{b}$ is subject to two key constraints. First, exactly $M$ MAs will be placed, which is enforced by
\begin{equation}
\vspace{-0.05in}
\sum\nolimits_{n=1}^{N}b_{n}=M.
\label{mbinb}
\end{equation}
Second, to mitigate mutual coupling between any pair of MAs, the distance between their positions must be at least the minimum separation $d_{\min}$ \cite{ycjin}, i.e.,
\begin{equation}
 \|\mathbf{p}_n-\mathbf{p}_{n'}\|\ge b_n b_{n'}d_{\text{min}},~\forall n,n'\in\mathcal{N}, ~n\ne n',
\label{mbinc}
\end{equation}
where $\mathcal{N} \triangleq \{1, 2, \dots, N\}$.

Let ${\mathbf{h}}_k \triangleq [h_{k1},{h}_{k2},\dots,{h}_{kN}]^T\in\mathbb{C}^{N}$ denote the channel between all $N$ CPs and the $k$-th UE. The received signal of the $k$-th UE is given by
\begin{equation}
\vspace{-0.1in}
y_{k} ={\mathbf{h}_{k}(\mathbf{b})}^{H}\mathbf{w}_{k}s_k+\sum_{l=1,l\ne k}^{K}{\mathbf{h}_{k}(\mathbf{b})}^{H}\mathbf{w}_{l}s_l + n_k,
\label{equ13}
\end{equation}
where $\mathbf{h}_k(\mathbf{b})\in \mathbb{C}^M$ denotes the channel between the selected CPs and the $k$-th UE, $\mathbf{w}_{k} \in \mathbb{C}^{M}$ represents the corresponding transmit beamformer, and $n_k$ represents the AWGN with power $\sigma_{k}^2$. The sum rate of all $K$ UEs is then expressed as
\begin{equation}
R^{\text{MA}}(\mathbf{b},\mathbf{w};\mathbf{h})=\sum_{k=1}^{K}\log_2\left(1 + \frac{| {\mathbf{h}_{k}(\mathbf{b})}^{H}\mathbf{w}_{k}|^{2}}{\sum\limits_{l=1,l\ne k}^{K}| \mathbf{h}_{k}^{H}{(\mathbf{b})}^{H}\mathbf{w}_{l} | ^{2}+\sigma_k^2}\right),
\label{equ14}
\end{equation}
where $\mathbf{w}=[\mathbf{w}_1^T,\mathbf{w}_2^T,\dots,\mathbf{w}_{K}^T]^T\in\mathbb{C}^{KM}$, and $\mathbf{h}=[\mathbf{h}_1^T,\mathbf{h}_2^T,\dots,\mathbf{h}_K^T,\mathbf{p}_1^T,\mathbf{p}_2^T,\dots,\mathbf{p}_N^T]^T\in\mathbb{C}^{KN+2N}$. We aim to maximize $R^{\text{MA}}$ with respect to $\bf{b}$ and $\mathbf{w}$, which is expressed as
\begin{subequations} 
\vspace{-0.2in}
\label{MAoptim}
\begin{align} 
\text{(P7):}&\max _{\mathbf{b}\in\{0,1\}^{N},\mathbf{w}\in \mathbb{C}^{KM}}& &R^{\text{MA}}(\mathbf{b},\mathbf{w};\mathbf{h})&
\label{MAobj} \\
&~~~~~~~~~~\text{s.t.}&  &\text{(\ref{mbinb}), (\ref{mbinc})},& \nonumber \\ 
&&  &\sum_{k=1}^{K}\left\|\mathbf{w}_{k}\right\|^{2} \leq P_{\text{max}},\label{mbind}&
\end{align}
\end{subequations}
where $P_{\text{max}}$ is the maximum transmit power of the BS. The problem (P7) also shares the same form with (P1), with $N_\text{b}=N$, $N_\text{w}=KM$, ${N_{\text{h}}}=KN+2N$, $\Omega=M^2-M+1$, and $\Psi=1$. Using \emph{Property 1}, the problem (P7) can be reformulated as 
\begin{subequations}
\vspace{-0.2in}
\label{MAoptim2}
\begin{align} 
\text{(P8):}&\max _{\mathcal{A}\in2^{\mathcal{N}},\mathbf{w}\in \mathbb{C}^{KM}}& &R_{\text{MA}}(\mathbf{b},\mathbf{w};\mathbf{h})&
\label{obj} \\
&~~~~~~~~\text{s.t.}& &\text{(\ref{mbinb})},\text{ (\ref{mbinc})},\text{ (\ref{mbind}),}& \nonumber\\
&& &{b_{n}}=\left\{\begin{array}{ll}
1,~~\forall n\in \mathcal{A}, \\
0,~~\forall n\notin \mathcal{A}.
\label{mdisb}\end{array}\right. 
\end{align}
\end{subequations}

Let $\mathcal{F}^{\text{MA}}(\cdot):\mathbb{C}^{KN+2N}\to 2^{\mathcal{N}}\times \mathbb{C}^{KM}$ represent the to-be-designed joint antenna positioning and beamforming scheme, (P8) can be further represented in the form of DL of (P3) as
\begin{subequations}
\vspace{-0.1in}
\begin{align}
\text{(P9):} ~&\max _{\mathcal{F}^{\text{MA}}(\cdot)} & &\mathbb{E}_{\mathbf{h}}[R^\text{MA}(\mathbf{b},\mathbf{w};\mathbf{h})]& \\
&~~\text{s.t.}& &\text{(\ref{mbinb}), (\ref{mbinc}), (\ref{mbind}), (\ref{mdisb})},& \nonumber \\
&& &\{\mathcal{A},\mathbf{w}\}=\mathcal{F}^{\text{MA}}(\mathbf{h}). \label{MALEARN}
\end{align}
\end{subequations}

\section{Proposed General DL Framework}

\begin{figure}[!t]
\centering
\includegraphics[width=0.4\textwidth]{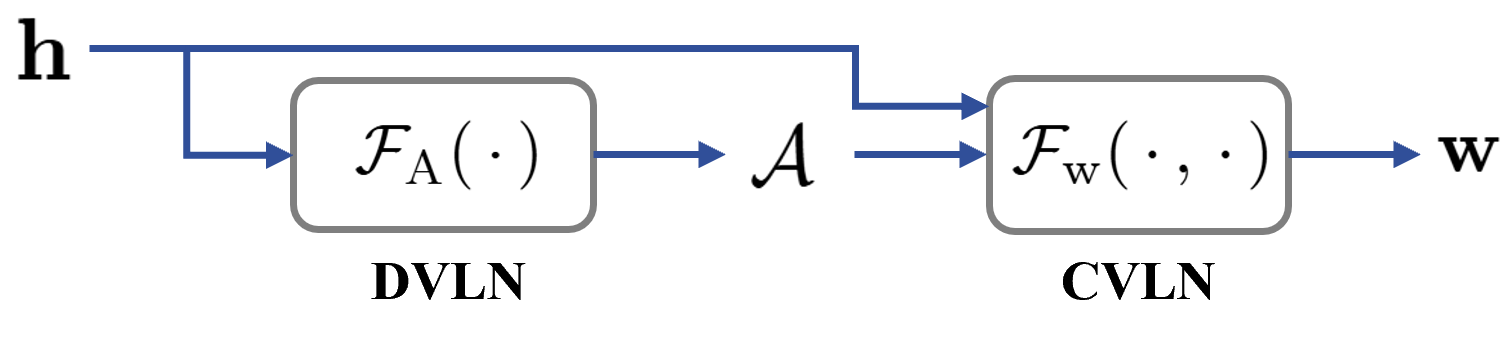}
\vspace{-0.15in}
\caption{The overall architecture of the proposed DL framework.}
\vspace{-0.15in}
\label{2stageFIG}
\end{figure}

In this section, we propose a general DL framework to learn the desired mapping $\mathcal{F}(\cdot):\mathbb{C}^{{N_{\text{h}}}}\to 2^{\mathcal{N}_{\text{b}}}\times \mathbb{C}^{N_{\text{w}}}$ in (P3). The proposed framework consists of two distinct models: a DVLN for discrete outputs and a CVLN for continuous outputs. These models are cascaded and trained jointly to provide near-optimal solutions while satisfying the constraints at the same time, as shown in Fig.~\ref{2stageFIG}. Specifically, the DVLN $\mathcal{F}_{\text{A}}(\cdot):\mathbb{C}^{{N_{\text{h}}}}\to2^{\mathcal{N}_{\text{b}}}$ produces the support set $\mathcal{A}$ of the discrete variable $\mathbf{b}$ by $\mathcal{A}=\mathcal{F}_{\text{A}}(\bf{h})$. Subsequently, the continuous variable $\mathbf{w}$ is produced by $\mathbf{w}=\mathcal{F}_{\text{w}}(\mathcal{A},\mathbf{h})$, where $\mathcal{F}_{\text{w}}(\cdot,\cdot):2^{\mathcal{N}_{\text{b}}}\times\mathbb{C}^{{N_{\text{h}}}}\to\mathbb{C}^{N_{\text{w}}}$ denotes the CVLN.

To tackle the zero-gradient problem of discrete outputs, we first model the entries in $\mathcal{A}$ as random variables and aim to learn their conditional probability distribution given the system parameter $\mathbf{h}$, i.e., $p(\mathcal{A}|\mathbf{h})$. Noticing that the constraints in (\ref{binb}) are coupled, which means that each element in $\mathcal{A}$ depends on other elements, we express $p\left(\mathcal{A}|\mathbf{h}\right)$ in its factorized form\footnote{\rev{A similar sequential factorization was independently proposed in~\cite{LSJWideband} using a sequential graph neural network (SGNN). Our key distinctions are constraint masking and direct utility optimization without softmax relaxation.}}
\begin{equation}
p(\mathcal{A}|{\mathbf{h}})= \prod_{t=1}^{T} p(a_{t} \mid \mathcal{A}_{t-1}, {\mathbf{h}}),
\label{factorize}
\end{equation}
where $a_t\in \mathcal{A}$ denotes the selected element of $\mathcal{A}$ at the $t$-th step, $\mathcal{A}_{t-1}\triangleq\{a_1,a_2,\dots,a_{t-1}\}$ represents the elements selected up to the ($t-1$)-th step with $\mathcal{A}_0\triangleq\emptyset$, and $T$ denotes the cardinality of $\mathcal{A}$. In some cases, the value of $T$ is exactly determined from the constraints in~\eqref{binb}. For example, in \textbf{Example 2} of Section~\uppercase\expandafter{\romannumeral2}-C, $T=M$, which is governed by~\eqref{mbinb}. On the other hand, in \textbf{Example 1} of Section~\uppercase\expandafter{\romannumeral2}-C, the constraints \eqref{equ6} and \eqref{equ7} only provide an upper bound on the number of elements in $\mathcal{A}$. In this case, we need a mechanism that allows us to stop adding element to the set $\mathcal{A}$ before reaching the upper bound. To this end, we employ an end token in \rev{DL}, which will be elaborated later. To facilitate understanding, we focus on the case where we know the exact number of elements in $\mathcal{A}$ at this moment.    

It is observed from (\ref{factorize}) that this factorization enables a sequential construction of $\mathcal{A}$ over $T$ steps. At each step $t$, we calculate the probability of each possible value of $a_t$, i.e., $\{p(a_{t}=n \mid \mathcal{A}_{t-1}, {\mathbf{h}})\}_{n\in {\mathcal{N}_\text{b}\setminus \mathcal{A}_{t-1}}}$ where $\mathcal{N}_\text{b}\setminus \mathcal{A}_{t-1}$ denotes remaining candidates. The next element $a_t$ is then selected by probabilistic sampling or greedy selection based on $p(a_{t} \mid \mathcal{A}_{t-1}, {\mathbf{h}})$. 

\rev{It is worth noting that the number of decoding steps $T$ is bounded by small constants in both case studies  (e.g., $T \le \min \{KL_{\max}, LK_{\max}\}$ in CF systems and $T=M$ in MA-aided systems), so the sequential overhead does not scale with the overall network size. Furthermore, even when the problem involves multiple-level variables (e.g., quantized power allocation), the decoding depth $T$ increases by exactly one per variable. This is efficiently handled by the masking mechanism: once a specific level is selected, all other mutually exclusive candidates are immediately masked out, avoiding any combinatorial expansion.}

In the following subsections, we present the architectures of $\mathcal{F}_{\text{A}}(\cdot)$ and  $\mathcal{F}_{\text{w}}(\cdot,\cdot)$, and demonstrate that the proposed $\mathcal{F}_{\text{A}}(\cdot)$ effectively captures the non-SPSD property inherent in the discrete optimization problem.

\vspace{-0.1in}
\subsection{Design of $\mathcal{F}_{\text{A}}(\cdot)$}
\begin{figure}[!t]
\centering
\includegraphics[width=0.5\textwidth]{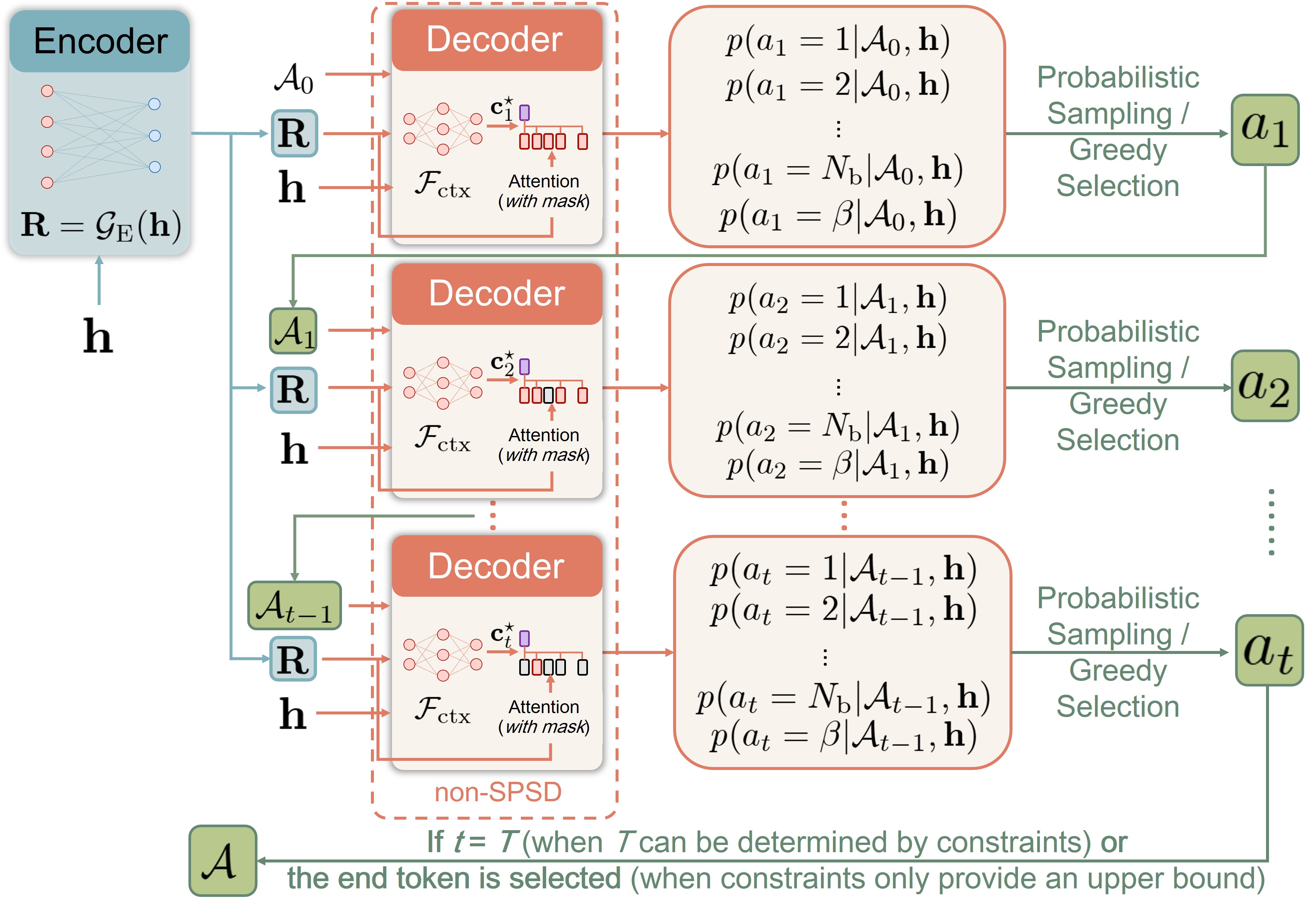}
\vspace{-0.1in}
\caption{Overview of the DVLN.}
\vspace{-0.15in}
\label{Fig2}
\vspace{-0.1in}
\end{figure}

To learn the conditional probability $p(\mathcal{A}|\mathbf{h})$ while exploiting~\eqref{factorize}, we propose a DVLN with an encoder-decoder structure, where the encoder generates the embeddings of all $N_{\text{b}}$ elements in $\mathcal{N}_{\text{b}}$, and the decoder outputs the conditional probabilities \textit{sequentially} in $T$ steps. The whole process of the DVLN is summarized in Fig.~\ref{Fig2}, and the design details are introduced as follows.

\textit{1) Design of the encoder:} The encoder takes the system parameter $\mathbf{h}$ as the input, and outputs a $d_{\text{h}}\times N_{\text{b}}$ embedding:
\begin{equation}
\label{embed}
\mathbf{R}\triangleq[\mathbf{r}_1,\mathbf{r}_2,\dots,\mathbf{r}_{N_{\text{b}}}]= \mathcal{G}_\text{E}\left(\mathbf{h}\right),
\end{equation}
where $\mathbf{r}_{n}\in \mathbb{R}^{d_{\text{h}}}$ is a $d_{\text{h}}$-dimensional vector representing the embedding of the $n$-th element of $\mathbf{b}$, and $\mathcal{G}_\text{E}(\cdot):\mathbb{C}^{{N_{\text{h}}}}\to \mathbb{R}^{d_{\text{h}}\times N_{\text{b}}}$ denotes the encoding scheme.

The encoder network can be implemented using various
architectures, such as multilayer perceptron (MLPs),  GNNs, or transformers. In Sections \uppercase\expandafter{\romannumeral5} and \uppercase\expandafter{\romannumeral6}, we will design custom architectures of $\mathcal{G}_\text{E}(\cdot)$ for the two examples presented in Section \uppercase\expandafter{\romannumeral2}-C.

\textit{2) Design of the decoder:} As described earlier, the decoder \emph{sequentially} constructs the support set $\mathcal{A}$ over $T$ steps. At the $t$-step, it selects an element $a_t$ based on $\{p(a_t = n |\mathcal{A}_{t-1}, \mathbf{h})\}_{n\in {\mathcal{N}_\text{b}\setminus \mathcal{A}_{t-1}}}$. These probabilities are predicted via an attention mechanism that computes the relevance between the embeddings in $\mathbf{R}$ and a context representation of the current system state $\left(\mathcal{A}_{t-1},\mathbf{h}\right)$. In particular, we introduce a \emph{context embedding} network to capture the current system state:
\begin{equation}
\mathbf{c}^{\star}_{t}={\mathcal{F}_{\text{ctx}}(\mathcal{A}_{t-1}, \mathbf{h},\mathbf{R}), } 
\label{con_embed}
\end{equation}
where $\mathbf{c}^{\star}_{t}\in \mathbb{R}^{d_{\text{h}}}$ denotes a $d_\text{h}$-dimensional context vector of the system state at step $t$, and $\mathcal{F}_{\text{ctx}}(\cdot,\cdot,\cdot)$ is a function parameterized by a NN. \rev{This \textit{dynamic context embedding} $\mathbf{c}_t^{\star}$ effectively reflects the current system state while simultaneously captures the non-SPSD property, as will be elaborated in detail in Section~III-C.} Then, given $\mathbf{c}_{t}^{\star}$, a compatibility score for each candidate is computed as \cite{AAN,kool2018attention}:
\begin{equation}
u_{n,t} =
\text{ATT}(\mathbf{r}_{n},\mathbf{c}^{\star}_{t}),~~\forall n\in {\mathcal{N}_\text{b}\setminus \mathcal{A}_{t-1}},
\label{nomaskATT}
\end{equation}
where $\text{ATT}(\cdot,\cdot)$ denotes a general attention-based NN \cite{AAN,kool2018attention} that quantifies the relevance between $\mathbf{r}_{n}$ and $\mathbf{c}^{\star}_{t}$. Finally, the conditional probability is then obtained via a softmax normalization over the compatibility scores:
\begin{equation}
\label{Pointeroutput}
p(a_t=n \mid \mathcal{A}_{t-1}, \mathbf{h})=\frac{e^{{u}_{n,t}}}{\sum_{n'=1}^{N_{\text{b}}} e^{{u}_{n',t}}},\quad\forall n\in {\mathcal{N}_\text{b}\setminus \mathcal{A}_{t-1}}.
\end{equation}

However, the attention mechanism described above does not inherently respect the discrete constraints in~\eqref{binb}, as it may assign non-zero probabilities to candidates that would render the solution infeasible. To guarantee feasibility at every decoding step, we explicitly mask out any candidate that violates the constraints if added to the current support set. Specifically, we modify the attention scores in~\eqref{nomaskATT} by:
\begin{equation}
\begin{split}
u_{n,t} &= \begin{cases}
\text{ATT}(\mathbf{r}_{n},\mathbf{c}^{\star}_{t}) & ,~\forall n\in {\mathcal{N}_\text{b}\setminus \mathcal{A}_{t-1}}\setminus\tilde{\mathcal{N}}_{\text{b},t}, \\
-\infty &,~\forall n\in \tilde{\mathcal{N}}_{\text{b},t},
\end{cases} \\
\end{split}
\label{Compati}
\end{equation}
where $\tilde{\mathcal{N}}_{\text{b},t} \subseteq \mathcal{N}_{\text{b}}$ \rev{is defined as all candidates whose addition to the current partial solution $\mathcal{A}_{t-1}$ would violate any constraint in (1b),} which is formally defined as
\begin{equation}
\begin{aligned}
\tilde{\mathcal{N}}_{\text{b},t} \triangleq & \left\{n \in \mathcal{N}_{\text{b}}\setminus \right.  
\mathcal{A}_{t-1}|\\
&\left. \exists m \in \{1,...,\Omega\}, ~f_m \left(\mathbf{b}(\mathcal{A}_{t-1}\cup\{n\})\right) > 0 \right\},
\end{aligned}
\label{MaskSTG}
\vspace{-0.05in}
\end{equation}
with $\mathbf{b}(A_{t-1}\cup\{n\})$ representing the discrete variable $\mathbf{b}$ with support set $\mathcal{A}_{t-1} \cup \{n\}$. It can be seen from \eqref{Pointeroutput}, \eqref{Compati} and \eqref{MaskSTG} that for all elements violating the constraints in (\ref{binb}), the corresponding conditional probability $p(a_t=n|\mathcal{A}_{t-1},\mathbf{h})$ will approach zero. \rev{Therefore, the decoding process only selects candidates that keep the partial solution extendable to a fully feasible one. By induction, the final solution $\mathcal{A}$ is guaranteed to satisfy all discrete constraints.}
In cases where the constraints~\eqref{binb} only provide an upper bound to $T$ (in the worst case, we can always set the upper bound as $N_{\text{b}}$), we need the capability to stop adding element to $\mathcal{A}$ before reaching the upper bound. In such cases, we introduce an end token $\beta$. Specifically, at each decoding step $t$, we augment the candidate set ${\mathcal{N}_\text{b}\setminus \mathcal{A}_{t-1}}$ as $\mathcal{N}_{\text{b}}\setminus\mathcal{A}_{t-1}\cup\{\beta\}$. The attention score associated with $\beta$ is defined in the form of~\eqref{Compati} but with $u_{\beta,t}=\text{ATT}\left(\mathbf{r}_{\beta},\mathbf{c}_{t}^{\star}\right)$, where $\mathbf{r}_{\beta}\in\mathbb{R}^{d_{\text{h}}}$ is a trainable vector. If $\beta$ is selected at step $t$, the decoding process will stop and the final support set is fixed as $\mathcal{A}=\mathcal{A}_{t-1}$. 

\rev{\textit{Remark 1: }It is worth noting that the masking in \eqref{Compati} and \eqref{MaskSTG} is a constructive constraint-enforcement mechanism rather than a post-processing heuristic, and that the \emph{sequential} decisions are trained with the global utility objective while respecting all coupled constraints via the masking mechanism.}

\subsection{Design of $\mathcal{F}_{\text{w}}(\cdot,\cdot)$}
The CVLN $\mathcal{F}_{\text{w}}(\cdot,\cdot)$ takes $\mathcal{A}$ and $\mathbf{h}$ as input, and output the continuous variable $\mathbf{w}$. While numerous methods have been developed to learn such continuous mappings \cite{9505260,MDGNN,10970427,HPEtransformer,9941253,10707363}, the specific architecture of $\mathcal{F}_{\text{w}}(\cdot,\cdot)$ usually depends on the problem structure, and the case-specific implementation will be given in Sections \uppercase\expandafter{\romannumeral5} and \uppercase\expandafter{\romannumeral6}, respectively.


\vspace{-0.1in}
\subsection{Resolving the Non-SPSD Challenge in Discrete Optimization}

In this subsection, we formalize the non-SPSD property of (P3) and demonstrate that the proposed framework is explicitly designed to capture this property. The non-SPSD property characterizes scenarios in which identical environmental conditions may correspond to significantly different solutions, which is a phenomenon commonly observed in mixed-discrete wireless resource allocation problems.

\textit{Property 2 (Non-SPSD Property of (P3)): }Given that the system parameter of (P3) consists of $N_\text{b}$ blocks, i.e., $\mathbf{h} = [\mathbf{h}_1^T, \mathbf{h}_2^T, \dots, \mathbf{h}_{N_\text{b}}^T]^T$, proximity of parameter blocks does \textit{not} guarantee identical membership in $\mathcal{A}$. More precisely, for any given tolerance $\varepsilon \ge 0$, there exist instances of (P3) (i.e., choices of $\mathbf{h}$) and distinct indices $i$ and $j$ such that the corresponding blocks are $\varepsilon$-close, yet their membership status in $\mathcal{A}$ differs. Formally, non-SPSD property is defined as
\begin{equation}
\begin{aligned}
\exists i, j:~ & ||\mathbf{h}_i - \mathbf{h}_j|| \leq \varepsilon \text{ and } \nonumber\\
&\left( (i \in \mathcal{A} \text{ and } j \notin \mathcal{A}) \text{ or } (i \notin \mathcal{A} \text{ and } j \in \mathcal{A}) \right). \nonumber
\end{aligned}
\end{equation}

Many mixed-discrete wireless resource allocation problems possess this non-SPSD property, including the two examples elaborated in Section~II-C. \rev{For instance, in (P6), when the UE-AP pair $(k,l)$ and $(k',l')$ have identical channel conditions, i.e., $\mathbf{h}_{kl}=\mathbf{h}_{k'l'}$, their association solutions can be $(k,l)\in\mathcal{A}$ and $(k',l') \notin \mathcal{A}$, due to the strong interference between them as modeled in the rate function \eqref{equ9}. Similarly, in (P9), when CP~$n$ and CP~$n'$ have near-identical channel conditions and positions, i.e., $|h_{kn}-h_{kn'}|<\epsilon$ and $\|\mathbf{p}_n-\mathbf{p}_{n'}\|<\epsilon$, their positioning solutions can be $n \in \mathcal{A}$ and $n' \notin \mathcal{A}$, due to the coupled discrete constraints~\eqref{mbinc}.}
\rev{Crucially, thanks to the \emph{sequential decoding strategy} and the \emph{dynamic context embedding}, the proposed DVLN is capable of capturing the non-SPSD property on the discrete solutions, as stated below.}

\vspace{+0.1in}

\noindent \textbf{Proposition 1} \textit{(Non-SPSD Property of $\mathcal{F}_{\text{A}}(\cdot)$)}:  Let $\mathbf{h}=[\mathbf{h}_1^T, \mathbf{h}_2^T,..., \mathbf{h}_{N_\text{b}}^T]^T$. The DVLN $\mathcal{F}_{\text{A}}(\cdot)$ is capable of producing the output $\mathcal{A}$ such that $\left(i \in \mathcal{A} \text{ and } j \notin \mathcal{A}\right) \text{ or } \left(i \notin \mathcal{A} \text{ and } j \in \mathcal{A}\right) $ even if $\mathbf{h}_i=\mathbf{h}_j$.

\begin{proof}
When $\mathbf{h}_i=\mathbf{h}_j$, the encoder $\mathcal{G}_{\text{E}}(\cdot)$ in~\eqref{embed} outputs identical embeddings $\mathbf{r}_i=\mathbf{r}_j$, since the encoder cannot distinguish between identical inputs without a judicious design~\cite{LSJWideband}. We can see
from \eqref{Compati} that their attention scores are also the same, i.e., $u_{i,t}=u_{j,t}$ if both $i$ and $j$ belong to the candidate set $\tilde{\mathcal{N}}_{\text{b},t}$ at step $t$. However, since at most only one of them (say $i$) can be selected to be included in $\mathcal{A}_{t}$, this makes the context vector $\mathbf{c}_{t+1}^*$ at step $t+1$ different from the previous $\mathbf{c}_{t}^*$ at step $t$. Consequently, the attention score of the remaining $j$ will also change in the subsequent decoding steps.
This enables the capability of $\mathcal{F}_\text{A}(\cdot)$ to produce $\mathcal{A}$ such that $i\in\mathcal{A}$ while $j \notin \mathcal{A}$ (or vice versa). 
\end{proof}

\rev{The symmetry-breaking capability of autoregressive decoders due to sequential conditioning has been observed in combinatorial optimization, e.g., in POMO~\cite{kwon2020pomo}. Our work adopts a similar sequential decoding mechanism, but differs in that we focus on adapting this paradigm to wireless problems where non-SPSD arises from interference and coupled constraints, rather than from problem symmetry. A POMO-style multi-start strategy could be integrated as a future extension to further enhance performance.}

\rev{\textit{Remark 2: }While the sequential autoregressive decoding and attention mechanism share similarities with~\cite{bello2016neural,kool2018attention}, the proposed framework differs in its handling of mixed variables, coupled discrete constraints, unknown support set cardinality, and non-SPSD. These differences are essential for wireless resource allocation.}

\section{Proposed Training Algorithm}
During training, each element $a_t$ is obtained via probabilistic sampling from its conditional distribution, i.e., $a_{t}\sim p(a_{t}|\mathcal{A}_{t-1},\mathbf{h})$, to expand the scope of exploration. Then, based on the probabilistic modeling $p(\mathcal{A}|\mathbf{h})$, the joint training of (P3) can be formulated as
\begin{subequations}
\begin{align}
\text{(P10): }&\max_{\boldsymbol{\theta}_{\text{A}},\boldsymbol{\theta}_{\text{w}}}& &J(\boldsymbol{\theta}_{\text{A}},\boldsymbol{\theta}_{\text{w}})=\mathbb{E}_{\mathbf{h},\mathcal{A}\sim p(\mathcal{A}|\mathbf{h})}[U(\mathbf{b},\mathbf{w};\mathbf{h})], \tag{26} \label{DNNjointrain}
\end{align}
\end{subequations}
where $\boldsymbol{\theta}_{\text{A}}$ and $\boldsymbol{\theta}_{\text{w}}$ denote the trainable parameters of $\mathcal{F}_{\text{A}}(\cdot)$ and $\mathcal{F}_{\text{w}}(\cdot,\cdot)$, respectively, and the original constraints in (\ref{binb}) and (\ref{binc}) are enforced through the design of $\mathcal{F}_\text{A}(\cdot)$ and $\mathcal{F}_\text{w}(\cdot,\cdot)$, respectively. \rev{The joint training of DVLN and CVLN using~\eqref{DNNjointrain} implies that the DVLN learns to make discrete decisions specifically tailored to provide the
best possible foundation for the CVLN, rather than optimizing a surrogate metric in isolation.} Next, we show how to update $\boldsymbol{\theta}_\text{A}$ and $\boldsymbol{\theta}_\text{w}$.

\emph{1) Update of $\boldsymbol{\theta}_{\text{A}}$}: Since $\mathcal{A}$ is sampled according to $p(\mathcal{A}|\mathbf{h})$
during training, $U(\mathbf{b}, \mathbf{w}; \mathbf{h})$ is non-differentiable with respect to $\boldsymbol{\theta}_{\text{A}}$. However, since $p(\mathcal{A}|\mathbf{h})$ itself is differentiable with
respect to $\boldsymbol{\theta}_{\text{A}}$, we equivalently express the gradient of $J(\boldsymbol{\theta}_{\text{A}}, \boldsymbol{\theta}_{\text{w}})$ with
respect to $\boldsymbol{\theta}_{\text{A}}$ as
\begin{equation}
\begin{aligned}
&\nabla_{\boldsymbol{\theta}_{\text{A}}} J(\mathbf{\boldsymbol{\theta}}_{\text{A}},\boldsymbol{\theta}_{\text{w}}) = \mathbb{E}_{\mathbf{h}}\left[\sum_{\mathcal{A}\sim p(\mathcal{A}\mid \mathbf{h})} U(\mathbf{b},\mathbf{w};\mathbf{h}) \nabla _{\boldsymbol{\theta}_{\text{A}}}p(\mathcal{A} \mid \mathbf{h})\right] \\
&= \mathbb{E}_{\mathbf{h}}\left[\sum_{\mathcal{A}\sim p(\mathcal{A}\mid \mathbf{h})} U(\mathbf{b},\mathbf{w};\mathbf{h}) p(\mathcal{A}|\mathbf{h})\nabla _{\boldsymbol{\theta}_{\text{A}}}\log p(\mathcal{A} \mid \mathbf{h})\right] \\
&=\mathbb{E}_{\mathbf{h},\mathcal{A}\sim p(\mathcal{A}\mid \mathbf{h})}\left[U(\mathbf{b},\mathbf{w};\mathbf{h})
\nabla_{\boldsymbol{\theta}_{\text{A}}} \log p(\mathcal{A}\mid \mathbf{h})\right].
\end{aligned}
\label{pg}
\end{equation}
In practice, the expectation in \eqref{pg} is approximated via Monte Carlo sampling over mini-batches, which corresponds to the well-known policy gradient method. To reduce the variance of the gradient estimates, we replace $U(\mathbf{b},\mathbf{w};\mathbf{h})$ in (\ref{pg}) with $U(\mathbf{b},\mathbf{w};\mathbf{h})-\hat{U}(\mathbf{h})$,
where $\hat{U}(\cdot):\mathbb{C}^{N_{\text{h}}} \to\mathbb{R}$ serves as an estimate of the sum rate, also known as the critic-network \cite{ActorCritic}, whose NN-architecture requires a case-specific design. The parameters $\boldsymbol{\theta}_\text{A}$ are then updated by a mini-batch stochastic gradient ascent algorithm, which can be implemented with the Adam optimizer \cite{kingma2014adam}. Correspondingly, the trainable parameter of $\hat{U}(\cdot)$, denoted by $\boldsymbol{\theta}_\text{C}$, is updated by a mini-batch stochastic gradient descent algorithm with the following mean-square-error loss function:
\begin{equation}
{L_{\text{MSE}}(\boldsymbol{\theta}_\text{C})} = \mathbb{E}_{\mathbf{h},\mathcal{A}\sim p(\mathcal{A}|\mathbf{h})}\left[ \left(\hat{U}(\mathbf{h})-U(\mathbf{b},\mathbf{w};\mathbf{h})\right)^2\right].
\label{MSE}
\end{equation}

\rev{While the update of $\boldsymbol{\theta}_{\text{A}}$ is the standard policy-gradient method REINFORCE~\cite{williams1992simple} with a learned baseline, our contribution lies in adapting it to the mixed-discrete wireless setting through joint training with the CVLN, masking for constraint enforcement, a domain-aware critic network for variance reduction, and an end-token to learn the effective stopping point. In fact, the proposed DVLN is agnostic to the choice of the policy-gradient optimizer. Advanced methods such as PPO~\cite{schulman2017proximal} can also be used to train the DVLN. Empirically, we observe that PPO achieves a similar final sum rate to the standard REINFORCE in our case studies. We adopt REINFORCE for its simplicity and lower hyperparameter sensitivity, while noting that PPO can be seamlessly integrated if sample efficiency is a primary concern. We do not recommend off-policy methods like SAC-Discrete~\cite{christodoulou2019soft} in our setting, as the continuously updated CVLN makes the replay buffer inconsistent, and sample generation via forward passes is already inexpensive.}

\addtolength{\topmargin}{0.05in}
\begin{algorithm}[t]
\caption{Joint Training of $\mathcal{F}_{\text{A}}(\cdot)$ and $\mathcal{F}_{\text{w}}(\cdot,\cdot)$}
\begin{algorithmic}[1]
\STATE \textbf{Input} number of epochs $N^{\text{e}}$, steps per epoch $N^{\text{s}}$, and batch size $B$
\STATE \textbf{Initialize} $\boldsymbol{\theta}_{\text{A}}$, $\boldsymbol{\theta}_{\text{w}}$, and $\boldsymbol{\theta}_\text{C}$
\FOR{epoch$=1,2,\dots,N^{\text{e}}$}
\FOR{step$=1,2,\dots,N^{\text{s}}$}
\STATE Select a batch of input $\{\mathbf{h}\}$ from the training dataset
    \STATE Output $p(\mathcal{A}|\mathbf{h})$, $\mathcal{A}$, and $\mathbf{w}$ corresponding to each training sample $\mathbf{h}$ 
    \STATE $\boldsymbol{\theta}_{\text{w}} \gets \text{Adam}(\boldsymbol{\theta}_{\text{w}},-\nabla_{\boldsymbol{\theta}_{\text{w}}}J(\boldsymbol{\theta}_{\text{A}},\boldsymbol{\theta}_{\text{w}}))$ 
\STATE $\boldsymbol{\theta}_{\text{A}} \gets \text{Adam}(\boldsymbol{\theta}_{\text{A}}, -\nabla_{\boldsymbol{\theta}_{\text{A}} }J(\mathbf{\boldsymbol{\theta}}_{\text{A}},\boldsymbol{\theta}_{\text{w}}))$
\STATE $\boldsymbol{\theta}_\text{C} \gets \text{Adam}(\boldsymbol{\theta}_\text{C}, \nabla_{\boldsymbol{\theta}_\text{C}} L_\text{MSE}(\boldsymbol{\theta}_\text{C}))$
\ENDFOR
\ENDFOR
\end{algorithmic}
\label{algorithm1}
\end{algorithm}

\emph{2) Update of $\boldsymbol{\theta}_{\text{w}}$}: We assume that $U(\mathbf{b},\mathbf{w};\mathbf{h})$ is differentiable with respect to $\mathbf{w}$ and consequently with respect to $\boldsymbol{\theta}_\text{w}$ \cite{
9505260,MDGNN,10970427,HPEtransformer,9941253,10707363,PilotGNN,LSJWideband, 11021397}. Thus, we can compute the gradient of $J(\mathbf{\boldsymbol{\theta}}_{\text{A}},\mathbf{\boldsymbol{\theta}}_{\text{w}})$ with respect to $\boldsymbol{\theta}_{\text{w}}$ by
\begin{equation}
\nabla_{\boldsymbol{\theta}_{\text{w}}} J(\mathbf{\boldsymbol{\theta}}_{\text{A}},\boldsymbol{\theta}_{\text{w}}) = \mathbb{E}_{\mathbf{h},\mathcal{A}\sim p(\mathcal{A}\mid \mathbf{h})}\left[\nabla_{\boldsymbol{\theta}_{\text{w}}}U\left(\mathbf{b},\mathbf{w};\mathbf{h}\right)\right].
\label{wtrain}
\end{equation}
Correspondingly, $\boldsymbol{\theta}_\text{w}$ is updated by a mini-batch stochastic gradient ascent algorithm using the Adam optimizer.

The joint training algorithm is summarized in Algorithm~\ref{algorithm1}, where the negative gradient is adopted in line 7 and line 8 to set a gradient \emph{ascent} update in the Adam optimizer. After training, $\mathcal{F}_{\text{A}}(\cdot)$ and $\mathcal{F}_{\text{w}}(\cdot,\cdot)$ with parameters $\boldsymbol{\theta}_{\text{A}}$ and $\boldsymbol{\theta}_{\text{w}}$ are used to predict $\mathbf{b}$ and $\mathbf{w}$ given any system parameter $\mathbf{h}$. 

\begin{table*}[t]
\centering
\caption{Comparison of learning-based methods for wireless resource allocation}
\label{tab:comparison_methods}
\renewcommand{\arraystretch}{0}
\setlength{\tabcolsep}{5pt}

\begin{tabular}{
>{\raggedright\arraybackslash}p{0.14\textwidth}
>{\raggedright\arraybackslash}p{0.30\textwidth}
>{\raggedright\arraybackslash}p{0.23\textwidth}
>{\raggedright\arraybackslash}p{0.25\textwidth}
}
\toprule
\textbf{Method}
& \textbf{Feasibility for coupled discrete constraints}
& \textbf{Non-SPSD handling}
& \textbf{Joint training without relaxation} \\
\midrule
{\cite{park2025self}}
& N/A, no coupled discrete constraints
& \myxmark, not explicitly considered
& \myxmark, discrete variables only \\
\midrule
{SGNN \cite{LSJWideband}}
& \myxmark, usually requiring post-hoc projection or repair
& \mycmark, through sequential block generation
& \myxmark, due to softmax relaxation \\
\midrule
STE
& \myxmark, usually requiring post-hoc projection or repair
& \myxmark, parallel-output structure
& Partial, due to biased gradient approximation \\
\midrule
Gumbel-Softmax
& \myxmark, usually requiring post-hoc projection or repair
& \myxmark, parallel-output structure
& Partial, due to approximation error \\
\midrule
\textbf{Proposed}
& \mycmark, \textbf{through support-set reformulation and constraint-aware masking}
& \mycmark, \textbf{through non-SPSD-aware context embedding with dynamic attention}
& \mycmark, \textbf{through joint unsupervised training without relaxation} \\
\bottomrule
\end{tabular}
\end{table*}

\rev{\textit{Remark 3:} While policy-gradient training~\cite{park2025self} and sequential factorization~\cite{LSJWideband} have been used elsewhere, the proposed framework introduces three components specifically designed for the mixed-discrete problem (P1): 1) support-set reformulation with constraint-aware masking strategy for feasibility; 2) non-SPSD-aware context embedding with dynamic attention; and 3) joint unsupervised training of discrete and continuous variables without relaxation. To our knowledge, this is the first work to integrate these components for wireless resource allocation under discrete constraints, as illustrated in Table I.}

\textit{Remark 4:} To enhance the stability of the joint training, we can alternatively pretrain $\mathcal{F}_{\text{w}}(\cdot,\cdot)$ according to line 7 of Algorithm 1, where the input $\mathcal{A}$ corresponding to a given $\mathbf{h}$ can be randomly generated.

\section{Case Study 1: Joint UE-AP Association and Beamforming in CF Systems}

\begin{figure*}[t]
\centering
\includegraphics[width=0.75\textwidth]{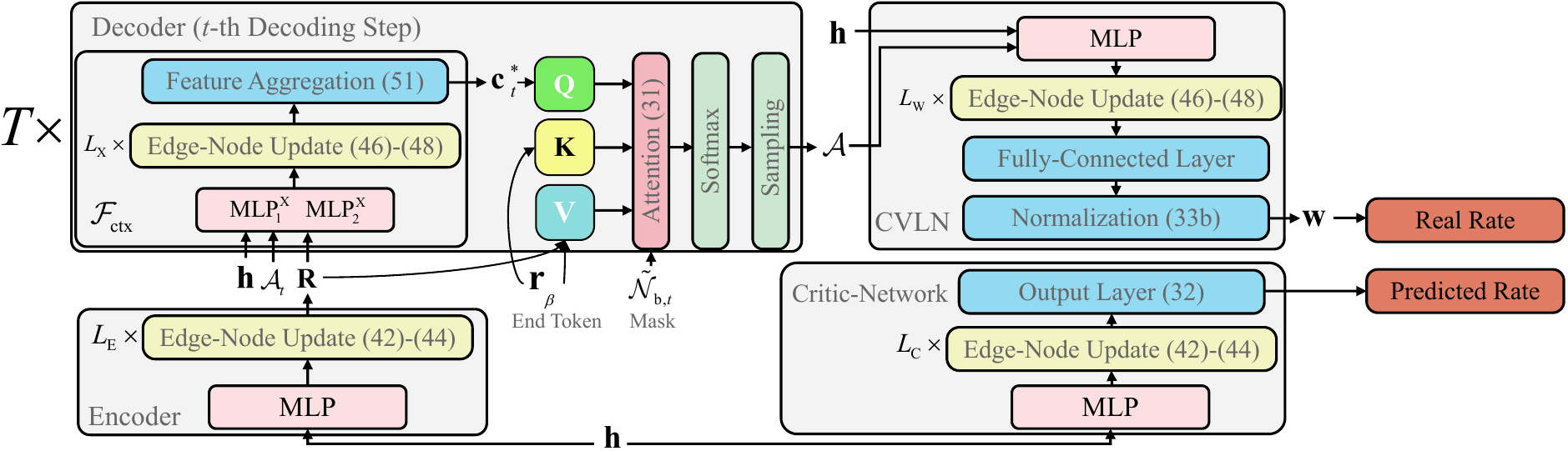}
\caption{The network architecture in Example 1.}
\label{CFArchitec}
\vspace{-0.2in}
\end{figure*}

In this section, we apply our proposed DL framework to solve a widely encountered wireless resource allocation problem: joint UE-AP association and beamforming in CF systems, as described in \textbf{Example 1} of Section \uppercase\expandafter{\romannumeral2}. We first detail the design of $\mathcal{F}_{\text{A}}(\cdot)$ and $\mathcal{F}_{\text{w}}(\cdot,\cdot)$, and then present simulations to demonstrate the superiority of the proposed DL framework compared with existing approaches. The whole NN architecture of our proposed framework for this case study is illustrated in Fig.~\ref{CFArchitec}.

\subsection{$\mathcal{F}_{\text{A}}(\cdot)$ Design Details}
\textit{1) Design of the encoder:} We need to design the encoder $\mathcal{G}_\text{E}(\cdot)$ as shown in (\ref{embed}), which takes $\mathbf{h} = [\mathbf{h}^T_{11}, \mathbf{h}^T_{12},\dots,\mathbf{h}^T_{1L},\mathbf{h}^T_{21},\dots, \mathbf{h}^T_{KL}]^T$ as the input, and outputs the embeddings $\mathbf{R} =
[\mathbf{r}_{11},\mathbf{r}_{12},
\dots,\mathbf{r}_{1L},\mathbf{r}_{21}, \dots,\mathbf{r}_{KL}]$. To this end, we model the CF system as a graph, which includes two types of nodes, i.e., $K$ UE-nodes and $L$ AP-nodes, and there exists an edge between each UE-node and each AP-node. Thus, $\mathcal{G}_\text{E}(\cdot)$ can be designed as a GNN. Specifically, we adopt the edge-node GNN (ENGNN) \cite{engnn} for its ability to jointly process edge and node features, offering greater expressive power.
 
In the ENGNN implementation, we denote the $k$-th UE-node feature and $l$-th AP-node feature at the $l_{\text{E}}$-th ($l_{\text{E}}=0,1,\dots,L_{\text{E}}$) hidden layer as $\mathbf{f}^{[l_{\text{E}}]}_{\text{UE},k}$ and $\mathbf{f}^{[l_{\text{E}}]}_{\text{AP},l}$, and the feature of their corresponding edge as $\mathbf{e}^{[l_{\text{E}}]}_{kl}$. The initial edge features are obtained by
\begin{equation}
\mathbf{e}_{kl}^{[0]}=\text{MLP}_1^{\text{E}}(\mathbf{h}_{kl}), ~~\forall k\in\mathcal{K},l\in \mathcal{L},
\label{preprocessCFE}
\end{equation}
where $\text{MLP}^{\text{E}}_1$ is an MLP, and the node features $\mathbf{f}_{\text{AP},l}^{[0]}$ and $\mathbf{f}_{\text{UE},k}^{[0]}$ are initialized as zero. Then, the node and edge features are updated by the node and edge update strategy, which is detailed in Appendix A (provided in the supplementary materials). After $L_{\text{E}}$ update layers, the final edge features serves as the required embedding, i.e., $\mathbf{r}_{kl}=\mathbf{e}_{kl}^{[L_{\text{E}}]}$.

\textit{2) Design of the decoder:}
We design the \emph{context embedding} network $\mathcal{F}_{\text{ctx}}(\cdot,\cdot,\cdot)$ as shown in (\ref{con_embed}), which takes $\mathcal{A}_{t-1}$, $\mathbf{R}$, and $\mathbf{h}$ as the input, and outputs a context vector $\mathbf{c}_t^{\star}$. To extract the CF network-level features, it is natural to employ a GNN as the backbone of $\mathcal{F}_{\text{ctx}}(\cdot,\cdot,\cdot)$.  Moreover, to incorporate the structural information $\mathcal{A}_{t-1}$ into the NN, unlike in the encoder $\mathcal{G}_{\text{E}}(\cdot)$ where all edges are treated homogeneously, we distinguish between two types of edges. Specifically, we categorize edges as associated edges, i.e., $\forall(k,l)\in\mathcal{A}_{t-1}$, and candidate edges, i.e., $\forall(k,l)\notin\mathcal{A}_{t-1}$, and employ distinct learnable parameters to extract their features. The detailed update equations are provided in Appendix B in the supplementary materials.

After computing the context vector $\mathbf{c}^{\star}_{t}$, the compatibility score for each candidate edge is computed via \eqref{Compati}, where the function $\text{ATT}(\cdot,\cdot)$ is achieved via a cross-attention mechanism, which is expressed as
\begin{equation}
\text{ATT}(\mathbf{r}_{kl},\mathbf{c}^{\star}_{t})= C \cdot \tanh\left(\frac{\mathbf{q}_t^{T}\mathbf{k}_{kl}}{\sqrt{d_\text{h}}}\right),
\label{cfattention}
\end{equation}
where $\mathbf{q}_t=\mathbf{W}_{\text{Q}} {\mathbf{c}^{\star}_t}$ and $\mathbf{k}_{kl}=\mathbf{W}_{\text{K}} \mathbf{r}_{kl}$ with trainable matrices $\mathbf{W}_{\text{Q}},\mathbf{W}_{\text{K}}\in \mathbb{R}^{d_\text{h}\times d_\text{h}}$, and the compatibilities are clipped within $\left[-C,C\right]$ with a hyper-parameter $C$.

In this case study, the total cardinality $T$ of the support set $\mathcal{A}$ is upper bounded by $\min\{KL_{\max}, LK_{\max}\}$, according to constraints \eqref{equ6} and \eqref{equ7}. To allow the possibility of the model to terminate before reaching this upper bound, we introduce the end token $\beta$ described at the end of Section III-A.

\emph{3) Design of the critic-network:} The design details of the critic-network can be found in Appendix C of the supplementary material.

\vspace{-0.1in}
\subsection{$\mathcal{F}_{\text{w}}(\cdot,\cdot)$ Design Details}
We need to design $\mathcal{F}_{\text{w}}(\cdot,\cdot)$, which takes $\mathbf{h}$ and $\mathcal{A}$ as the input, and outputs the beamforming variable $\mathbf{w}$. 
In particular, we adopt an architecture nearly identical to that of $\mathcal{F}_{\text{ctx}}(\cdot,\cdot,\cdot)$, differing only in the first and last layers. Specifically, the first layer of $\mathcal{F}_{\text{w}}(\cdot,\cdot)$ can be adjusted from (4) in Appendix B, by removing $\mathbf{r}_{kl}$ from the input and replacing $\mathcal{A}_{t-1}$ with $\mathcal{A}$. The node features are initialized as zero. After $L_{\text{W}}$ layers of updates, the beamformer $\mathbf{w}$ is obtained by 

\begin{subequations}
\begin{align}
&[\Re\{\mathbf{w}_{kl}^T\},\Im\{{\mathbf{w}^T_{kl}}\}]^T=\text{FC}^{\text{W}}(\hat{\mathbf{e}}_{kl}^{[L_{\text{W}}]}), \\
&\mathbf{w}_{kl}\gets\frac{\sqrt{P_{\text{max}}}b_{kl}{\mathbf{w}}_{kl}}{\sqrt{\max\left\{\sum_{k'=1}^{K}b_{k'l}\|{\mathbf{w}}_{k'l}\|^{2},P_{\max}\right\}}},~\forall k\in\mathcal{K},l\in\mathcal{L}.\label{normCF}
\vspace{-0.2in}
\end{align}
\end{subequations}
where $\hat{\mathbf{e}}_{kl}^{[L_{\text{W}}]}$ denotes the edge feature between the $k$-th UE and the $l$-th AP at the final hidden layer, $\text{FC}^{\text{W}}(\cdot)$ denotes a single-layer fully connected network, and \eqref{normCF} implements a projection layer that enforces the per-AP power constraints in (\ref{ccon}).

\vspace{-0.1in}
\subsection{Numerical Results}
In this subsection, we provide numerical results to demonstrate the benefits of the proposed DL framework for joint UE-AP association and beamforming in CF systems.

\textit{1) Simulation setting, choices of hyper-parameters, and benchmarks:}
The considered CF system consists of $L=8$ APs and $K=20$ UEs. Each AP is equipped with $M=4$ antennas and each UE has a single antenna. The maximum number of serving UEs of each AP is set as $K_{\max}=6$, and each user is served with at most $L_{\max}=2$ APs. The APs and UEs are randomly distributed at a $500\text{ m}\times500\text{ m}$ square area. The large-scale fading coefficients are generated according to the path-loss model $30.5 +
36.7 \log_{10}D_{kl}+ F_{kl}$ in dB, where $D_{kl}$ is the distance in meters between the $k$-th UE and the $l$-th AP, and $F_{kl}\sim\mathcal{N}(0,4)$ denotes the shadow fading component. The small-scale channel is generated from uncorrelated
Rayleigh fading. The noise power of each UE is set as $\sigma^2_k=-100\text{ dBm}$.

The hyper-parameters of the proposed DL framework are summarized as follows. All MLPs in our designed NNs have $2$ hidden layers, each with a fully connected layer, a batch normalization layer, and a ReLU activation. The encoder $\mathcal{G}_{\text{E}}(\cdot)$ has $L_{\text{E}}=2$ update layers, and the dimension of the hidden representation is set as $d_{\text{h}}=128$. The context embedding network $\mathcal{F}_{\text{ctx}}(\cdot,\cdot,\cdot)$ has $L_{\text{X}}=2$ update layers. The NN-based $\mathcal{F}_{\text{w}}(\cdot,\cdot)$ has $L_{\text{W}}=2$ hidden layers with dimension of hidden representation $128$. The critic-network has $L_{\text{C}}=6$ update layers. The hyper-parameter $C$ in~\eqref{cfattention} is set as $8$. In joint training, the number of training samples and validation samples are $51200$ and $1024$, respectively. The number of training epochs is set as $100$ with a batch size of $1024$. The learning rate is set as $10^{-4}$. After training, the performance is evaluated on $1024$ test samples, and in the testing stage, $a_{t}$ is determined by a greedy selection, i.e., $a_t = \text{argmax}_{(k,l)\in{\mathcal{K}\times \mathcal{L}}}~{p(a_t=(k,l)|\mathcal{A}_{t-1}, \mathbf{h})}$. The implementation of all simulations (including the compared methods) is under the PyTorch version 2.1.2+cu11.8, operating on an NVIDIA GeForce GTX 4090 GPU.

\begin{figure*}[!t]
    \centering
    \begin{minipage}{0.3\textwidth}
        \centering
        \includegraphics[width=\linewidth]{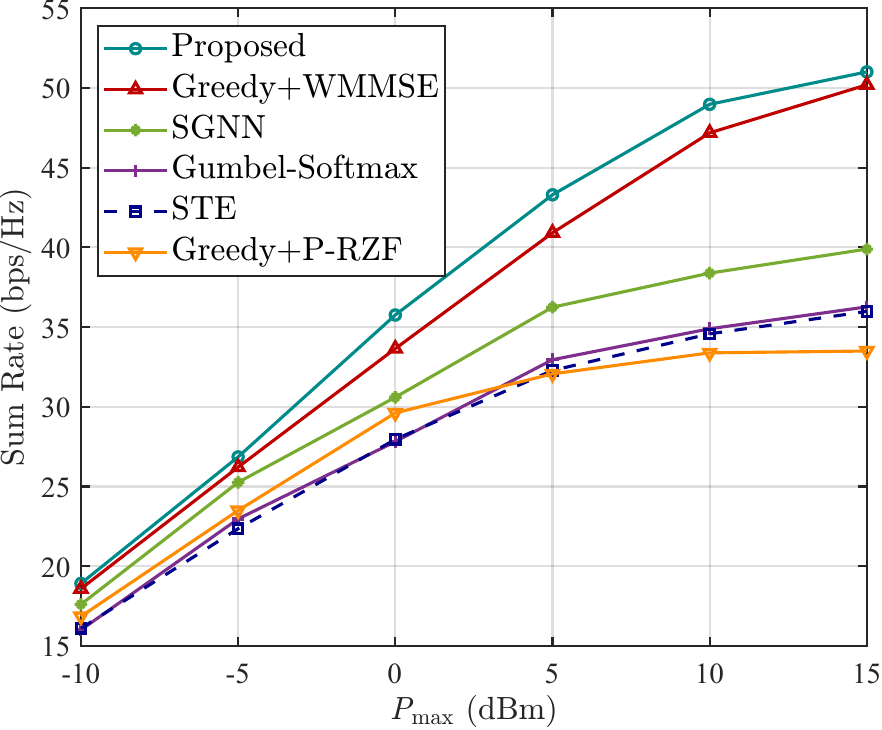}
        \caption{Sum rate performance comparison under different power budgets.}
        \label{fig:CFSumRate}
    \end{minipage}
    \hfill
    \begin{minipage}{0.32\textwidth}
        \centering
        \includegraphics[width=\linewidth]{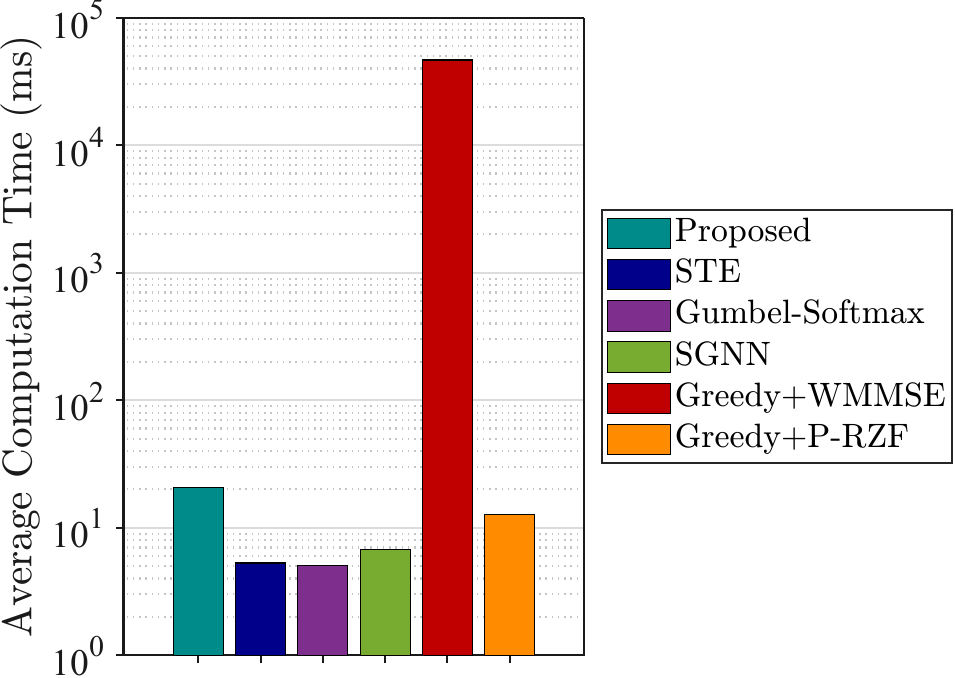}
        \caption{The average computation time comparison.}
        \label{fig:CFtime}
    \end{minipage}
    \hfill
    \begin{minipage}{0.32\textwidth}
        \centering
        \includegraphics[width=\linewidth]{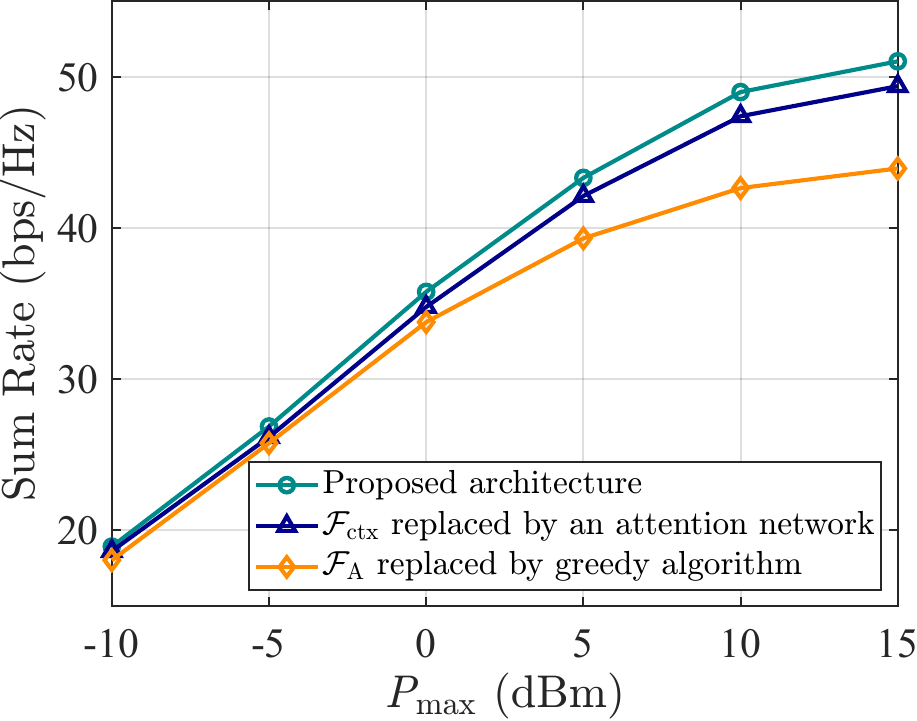}
        \caption{The sum rate performance with different context embedding network designs.}
        \label{fig:CFarchi}
    \end{minipage}
    \vspace{-0.2in}
\end{figure*}

For comparison, we provide the simulation results of the following benchmarks, including DL-based approaches and model-based approaches:
\begin{itemize}[left=0pt]
\item \textbf{Straight-Through-Estimator (STE): }This method employs an ENGNN to output discrete UE–AP association variables and continuous beamforming variables in parallel~\cite{PilotGNN}. The discrete association variables are handled by STE. To satisfy the association
constraints in~\eqref{equ6}, we deactivate extra links of those overloaded APs during the inference stage.
\item\textbf{Gumbel-Softmax: }This approach also utilizes an ENGNN to generate association and beamforming variables in parallel~\cite{11021397}. It further incorporates the Gumbel-Softmax reparameterization technique to approximate discrete sampling with differentiable operations~\cite{Gsoftmax}. Extra links to overloaded APs are deactivated during inference to satisfy the constraints in~\eqref{equ6}.
\rev{\item\textbf{SGNN: }To capture the non-SPSD property, this method utilizes a sequence of GNNs to learn the UE-AP association ~\cite{LSJWideband}. The zero-gradient issue is tackled via softmax approximation. Extra links to overloaded APs are deactivated during inference to satisfy the constraints in~\eqref{equ6}.}
\item \textbf{Greedy+WMMSE}: This approach first determines the UE-AP association via a greedy method~\cite{Heuristic}.  When a UE is accessed, it is served by $L_{\max}$ APs with the largest channel gains. If the selected AP already has $K_{\max}$ serving users, then the UE will be associated with the AP with next strongest channel gain. Given the association variable, the beamforming is designed by the the WMMSE algorithm \cite{WMMSE}, with the maximum number of iterations set to $50$ to ensure convergence.

\item \textbf{Greedy+P-RZF}: This method includes greedy-based UE-AP association and the partial regularized zero-forcing \cite{PRZF} beamforming.
\end{itemize}

\textit{2) Performance Evaluation:}
We first evaluate the sum rate performance of our proposed DL framework. In Fig.~\ref{fig:CFSumRate}, we show the sum rate performance against other baselines under different power budgets. It can be seen that the proposed DL framework consistently outperforms all baselines across all power levels. In particular, unlike \textit{STE}, which treats discrete variables as continuous and relies on the surrogate gradient, our approach directly handles discrete variables, thereby avoiding the gradient mismatch inherent in STE and achieving superior performance. A similar performance degradation occurs in \textit{Gumbel-Softmax}, due to the approximation error of discrete output by the Gumble-softmax function. \rev{Besides, these parallel-output baselines fail to capture non-SPSD, while our method succeeds, leading to further performance gap.} \rev{Furthermore, we can observe that SGNN, by capturing the non-SPSD property, outperforms parallel-output methods such as STE and Gumbel-Softmax. However, it still exhibits a noticeable performance gap compared with our proposed framework, mainly attributed to the approximation error induced by the softmax relaxation and the inability to enforce coupled discrete constraints.} Moreover, compared to model-based methods, the proposed DL framework outperforms than \emph{Greedy} heuristic, as the scheduling strategy becomes too complex to be handled by heuristic approaches.

Next, we compare the computational complexity of different methods, with results shown in Fig.~\ref{fig:CFtime}. We can see that our proposed DL framework achieves significantly faster inference speed than iterative optimization–based baselines \textit{Greedy+WMMSE} due to its computationally efficient feed-forward architecture.

Furthermore, we investigate the effect of changing the architecture design. In particular, in one implementation, we replace the proposed GNN-based context embedding network by an attention–based network, while all other components of the framework remain unchanged. Furthermore, we also experimented with replacing the $\mathcal{F}_\text{A}(\cdot)$ with the greedy algorithm, but keep the ENGNN-based beamforming as described in Subsection B. As shown in Fig.~\ref{fig:CFtime}, the proposed GNN-based design achieves higher sum rates than the attention-based approach, owing to its ability to effectively model edge features and thus offer greater expressive power \cite{engnn}. Nevertheless, both designs significantly outperform the \emph{Greedy} baseline, particularly at high-SNR regimes where interference management becomes critical.

\begin{figure*}[!t]
    \centering
        \begin{minipage}{0.31\textwidth}
        \centering
        \vspace{-0.1in}
        \includegraphics[width=\linewidth]{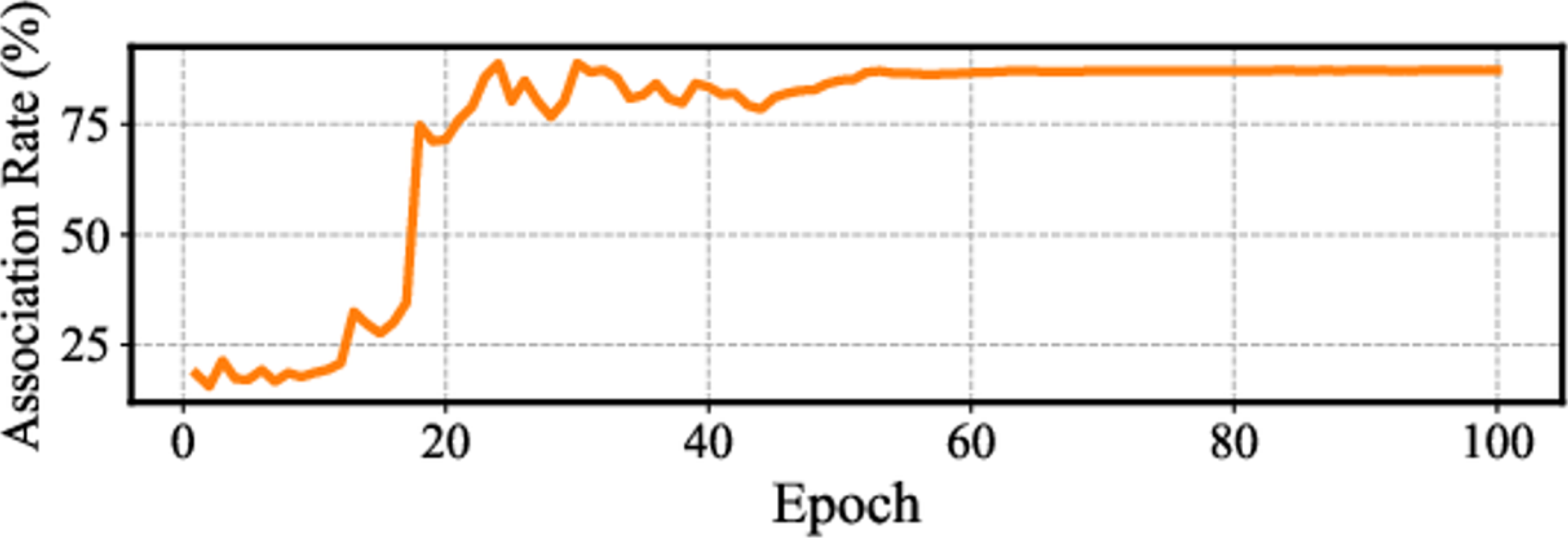}
        \caption{Association rate on the validation set versus training epoch.}
        \label{CFFig3}
    \end{minipage}
    \hfill
    \begin{minipage}{0.29\textwidth}
        \centering
        \vspace{-0.1in}
        \includegraphics[width=\linewidth]{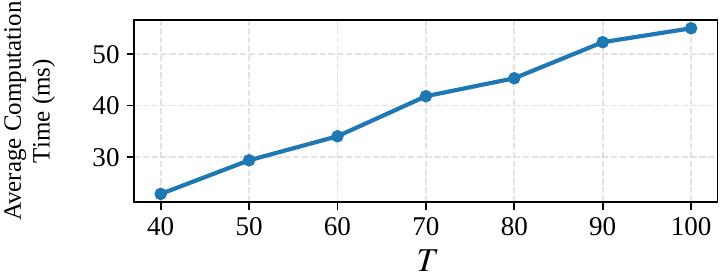}
        \caption{Average computation time under different $T$'s.}
        \label{fig:CF_time}
    \end{minipage}
    \hfill
    \begin{minipage}{0.35\textwidth}
        \centering
        \vspace{-0.1in}
        \includegraphics[width=\linewidth]{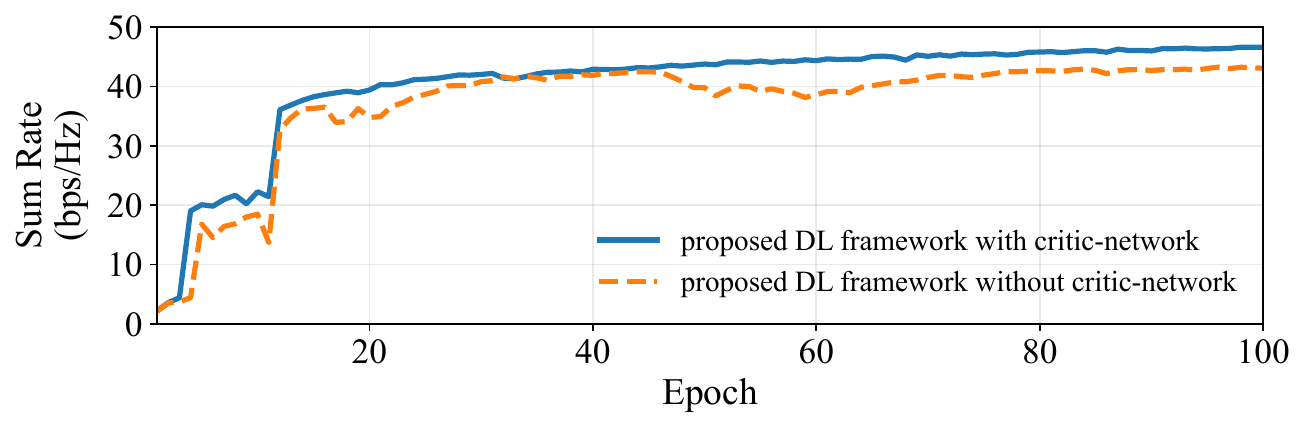}
        \caption{Impact of the critic-network.}
        \label{fig:CF_critic}
    \end{minipage}
    \hfill 
    \vspace{-0.2in}
\end{figure*}
We also examine the role of the end token in enabling the framework to learn an appropriate stopping point during the sequential UE-AP association process. In Fig.~\ref{CFFig3}, we set $P_{\max}=10\mathrm{dBm}$, and plot the average association rate on the validation set over the number of training epochs. The average association rate is defined as the obtained cardinality $T$ divided by its upper bound, i.e., ${T}/{\min\{LK_{\max},KL_{\max}\}}$. As shown Fig.~\ref{CFFig3}, the association rate is low initially, and significantly increases as training progresses and the NN explores more effective association strategies. This shows that the end token learns to determine an effective stopping point dynamically, rather than filling up the set $\mathcal{A}$ until the upper bound determined by the constraints.

\rev{To evaluate the scalability of the proposed approach with respect to $T$, Fig.~\ref{fig:CF_time} plots the inference time versus $T$, under the same system configuration as Fig.~\ref{fig:CFtime}. It can be observed that the inference time grows approximately linearly with $T$. Notably, even at $T = 100$, the total execution time remains below $60$ ms, which well satisfies the stringent latency requirements of practical wireless networks and offers a significant speedup over conventional iterative optimization-based methods.}

\rev{Moreover, we also compare the convergence behavior of the proposed framework with and without the critic-network in Fig.~\ref{fig:CF_critic} to validate the effectiveness of the critic-network. It can be observed that the use of the critic-network leads to faster and more stable convergence, demonstrating its effectiveness in stabilizing and accelerating the training process.}

\begin{figure}[t] 
    \centering
    \includegraphics[width=0.45\textwidth]{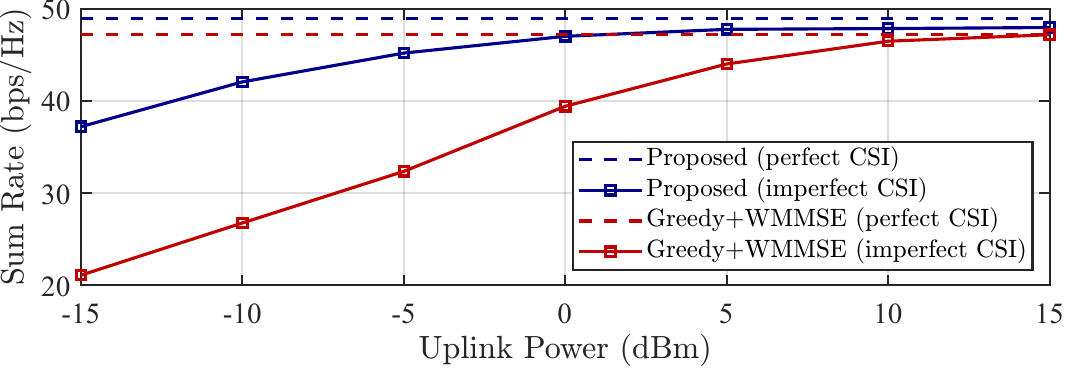}
    \vspace{-0.1in}
    \caption{Sum rate performance under imperfect CSI.}
    \label{fig:CF_imperfect_CSI}
\end{figure}

\rev{The above numerical results are based on the perfect channel state information (CSI). To further evaluate the robustness of the proposed DL framework under imperfect CSI, we adopt standard pilot-based minimum mean square error (MMSE) channel estimation~\cite{2020EmilScalable}, and use the estimated channels as inputs instead of ground-truth CSI. We set the pilot length to $20$ and the uplink noise power to $\sigma_{\text{ul}}^2=-100~\text{dBm}$, and vary the uplink transmit power. As observed in Fig.~\ref{fig:CF_imperfect_CSI}, when the uplink transmit power decreases and estimation errors become larger, the performance degradation of our DL framework is significantly smaller than that of \textit{Greedy+WMMSE}. This demonstrates the inherent robustness of DL-based methods to imperfect CSI.}

\section{Case Study 2: Joint Antenna Positioning and Beamforming in MA-aided Systems}
In this section, we employ our DL framework to address another wireless resource allocation problem: joint antenna positioning and beamforming for MA-aided systems, as described in \textbf{Example 2} of Section \uppercase\expandafter{\romannumeral2}.
\vspace{-0.12in}
\subsection{$\mathcal{F}_{\text{A}}(\cdot)$ Design Details}
\textit{1) Design of the encoder:} We need to design the encoder $\mathcal{G}_{\text{E}}(\cdot)$, which takes $\mathbf{h}=[\mathbf{h}_1^T,\mathbf{h}_2^T,\dots,\mathbf{h}_K^T,\mathbf{p}_1^T,\mathbf{p}_2^T,\dots,\mathbf{p}_{N}^{T}]^T$ as the input, and outputs the embeddings $\mathbf{R}=[\mathbf{r}_1,\mathbf{r}_2,\dots,\mathbf{r}_{N}]$. To this end, we model the MA-aided system as a graph, which includes two types of nodes, i.e., $K$ UE-nodes
and $N$ CP-nodes, and there exists an edge between each UE-node and each CP-node. 

We also adopt the ENGNN \cite{engnn} as the foundation architecture of $\mathcal{G}_{\text{E}}(\cdot)$, where the $k$-th UE-node feature and $n$-th CP-node feature of the $l_{\text{E}}$-th hidden layer are expressed as $\mathbf{f}^{[l_{\text{E}}]}_{\text{UE},k}$ and $\mathbf{f}^{[l_{\text{E}}]}_{\text{CP},n}$, and the corresponding edge feature is represented by $\mathbf{e}^{[l_{\text{E}}]}_{kn}$. The edge features and CP-node features are initialized as
\begin{subequations}
\begin{align}
&\mathbf{e}_{kn}^{[0]}=\text{MLP}^{\text{E}}_{1}\left(h_{kn}\right),~ \forall k \in \mathcal{K},n \in \mathcal{N},\label{MAEnPre} \\
&\mathbf{f}^{[0]}_{\text{CP},n}=\text{MLP}^{\text{E}}_{2}\left(\mathbf{p}_n\right),~~\forall n\in \mathcal{N},
\end{align}
\end{subequations}
where $\text{MLP}_1^{\text{E}}$ and $\text{MLP}_2^{\text{E}}$ are 2 MLPs, and the UE-node feature $\mathbf{f}^{[0]}_{\text{UE},k}$ is initialized as zero.
Using the ENGNN update equations, after $L_{\text{E}}$ hidden layers, the updated CP-node feature $\mathbf{f}^{[L_{\text{E}}]}_{\text{CP},n}$ serves as the required embedding $\mathbf{r}_{n}$.

\textit{2) Design of the decoder:} We design the context embedding function $\mathcal{F}_{\text{ctx}}(\cdot,\cdot,\cdot)$ which takes $\mathcal{A}_{t-1}$, $\mathbf{h}$, and $\mathbf{R}$ as the input, and outputs the context vector $\mathbf{c}^{\star}_t$. First, when $t>2$, we perform information aggregation to extract a preliminary graph-level feature, expressed as
\begin{equation}
\begin{aligned}
\mathbf{c}_{t}=&
\text{MLP}_2^{\text{X}}\left(\frac{1}{t-1}\sum\limits_{t'=1}^{t-1}\text{MLP}_1^{\text{X}}(\mathbf{r}_{a_{t'}}),\right. \\
& \left. \frac{1}{N}\sum_{n=1}^{N}\text{MLP}_4^{\text{X}}\left(\mathbf{p}_{n},\frac{1}{K}\sum_{k=1}^{K}\text{MLP}^{\text{X}}_{3}\left(h_{kn}\right)\right)\right),
\end{aligned}
\end{equation}
where $\text{MLP}^{\text{X}}_1$-$\text{MLP}^{\text{X}}_4$ are 4 MLPs. When $t=1$, as $\mathcal{A}_{t-1}=\emptyset$, we use a trainable parameter $\mathbf{r}^{\star} \in \mathbb{R}^{d_{\text{h}}}$ to replace the first input term of $\text{MLP}^{\text{X}}_2$. Next, we employ a multi-head attention (MHA) module to extract the deeper contextual features and refine the context vector. The MHA module calculates relevance scores between $\mathbf{c}_t$ and the embeddings of CPs, and integrates these score to enhance the representation of $\mathbf{c}_t$. The specific update equations are omitted due to space limitations.

Given $\mathbf{c}^{\star}_t$, the compatibility score for each CP is computed via \eqref{Compati}, where the function
$\text{ATT}(\cdot,\cdot)$ is achieved similar to~\eqref{cfattention}. In this case study, the total number of decoding steps can be determined as $T = M$ because of the constraint~\eqref{mbinb}.

\vspace{-0.1in}
\subsection{$\mathcal{F}_{\text{w}}(\cdot,\cdot)$ Design Details}
$\mathcal{F}_{\text{w}}(\cdot,\cdot)$ adopts an optimal solution structure to enable the NN to better learn the desired mapping. The design details are provided in Appendix D of the supplementary materials.

\vspace{-0.1in}
\subsection{Numerical Results}
\textit{1) Simulation setting and choices of hyper-parameters: }We consider an MA-aided system where the BS is equipped with $[4,9]$ MAs to serve $4$ single-antenna UEs. The size of the 2D rectangular transmit area is $2\lambda\times2\lambda$, where each side is uniformly sampled with $[5,8]$ points, resulting in $\{25,36,49,64\}$ CPs, and the wavelength $\lambda=60\text{ mm}$. The minimum distance between every two MAs is set as $d_{\text{min}}= 30\text{ mm}$. The distance between the $k$-th UE and the BS, denoted as $D_k$, is uniformly distributed within $[100,200]$ in meters. The noise power is set as $-100$ dBm.

We consider the field-response channel model\cite{10286328}, where the channel between the $n$-th CP and the $k$-th UE is expressed as
\begin{align}
{h}_{kn}=\mathbf{1}_{L_\text{p}}^{T} \boldsymbol{\Sigma}_{k} \mathbf{g}_{kn},~~\forall k \in \mathcal{K},~n \in \mathcal{N},
\end{align}
where $\mathbf{1}_{L_\text{p}}\in \mathbb{R}^{L_{\text{p}}}$ is the all-one field response vector (FRV), and $L_\text{p}=16$ denotes the path number.  The diagonal matrix $\boldsymbol{\Sigma}_{k}\triangleq\operatorname{diag}\left\{\left[\eta_{k,1},\eta_{k,2}, \dots, \eta_{k,L_{\text{p}}}\right]^{T}\right\}$, where $\eta_{k,l_\text{p}}$ denotes the response coefficient of the $l_{\text{p}}$-th path to the $k$-th UE, and it is independently and identically distributed as $\mathcal{C N}\left(0, L_{0} D_{k}^{-\alpha}\right)$ with $L_0=34.5$ dB and $\alpha=3.67$, respectively. Moreover, $\mathbf{g}_{kn}$ denotes the transmit FRV between the $n$-th CP and the $k$-th UE, which is given by
\begin{align}
    \mathbf{g}_{kn}=\left[e^{j \rho_{k,1}^{n}}, e^{j \rho_{k,2}^{n}},\cdots, e^{j \rho_{k,L_\text{p}}^{n}}\right]^{T},
\end{align}
where $\rho_{k,l_\text{p}}^{n}={2 \pi}/{\lambda}[\left(x_{n}-x_{1}\right) \cos \theta_{k, l_{\text{p}}} \sin \phi_{k, l_{\text{p}}}+(y_n-y_1)\sin\theta_{k,l_{\text{p}}}]$ represents the phase difference between the $n$-th CP and the first CP at the $l_\text{p}$-th path, and $\theta_{k, l_{\text{p}}}$ and $\phi_{k, l_{\text{p}}}$ represent the elevation angle of departure (AoD) and the azimuth AoD of the $l_\text{p}$-th path.
The probability density function of AoDs is $f_{\text{AoD}}\left({\theta}_{k, l_{\text{p}}}, \phi_{k, l_{\text{p}}}\right)={\cos {\theta}_{k, l_{\text{p}}}}/{2 \pi},~{\theta}_{k, l_{\text{p}}} \in\left[-{\pi}/{2},{\pi}/{2}\right],~\phi_{k, l_{\text{p}}} \in\left[-{\pi}/{2},{\pi}/{2}\right]$.

For the proposed NN model, all the MLPs are implemented by 2 linear layers, each followed by a ReLU activation function. The encoder $\mathcal{G}_{\text{E}}(\cdot)$ has $L_\text{E}=3$ layers, with $d_\text{h}=128$. The hyper-parameter is set to $C = 8$. The critic-network has $L_{\text{C}}=6$ update layers. Moreover, $\mathcal{F}_{\text{w}}(\cdot,\cdot)$ has $L_{\text{W}}=3$ hidden layers, each with a hidden dimension of $64$.  In the training procedure, the number of epochs is set to $100$, where each epoch consists of $50$ mini-batches with a batch size of $1024$. A learning rate $10^{-4}$ is adopted to update the trainable parameters through Algorithm \ref{algorithm1} using the Adam optimizer. The GPU and environment for training and testing is the same as in Section \uppercase\expandafter{\romannumeral5}.

\textit{2) Benchmarks:}
We consider four baseline methods for comparison:
\begin{itemize}[left=0pt]
\item \textbf{Random+WMMSE:} This method involves randomly selecting $M$ CPs. If the selected CPs violate the constraints in \eqref{mbinc}, the selection is repeated until the constraints are satisfied. Subsequently, with the MA positions determined, the beamformer is computed using the WMMSE algorithm\cite{WMMSE}.
\item \textbf{Greedy+WMMSE:} For this approach, the average channel gain of each CP with respect to all $K$ UEs is first calculated and treated as the equivalent channel gain for that CP. The CPs exhibiting the strongest channel gains are then selected iteratively over $M$ steps, with CPs violating (\ref{mbinc}) masked in each step. Following the MA positioning, the beamformer is obtained via the WMMSE algorithm.
\item \textbf{Greedy+ZF:} This method combines greedy algorithm for determining the MA position, and the zero-forcing beamforming~\cite{ZF}.
\item \textbf{FP-C:} We modify an approach that jointly optimizes continuous antenna positioning and beamforming for MA-aided systems \cite{C_FP_MA}. To obtain discrete antenna positions that satisfy the discrete constraints in \eqref{mbinc}, in each iteration of FP-C, we project the position of each MA to its nearest discrete CP that satisfies \eqref{mbinc} sequentially.
\end{itemize}
To the best of our knowledge, existing DL approaches cannot effectively handle the distance constraints in (\ref{mbinc}); therefore, we do not include DL-based benchmarks.

\begin{figure*}[!t]
\centering
\subfigure[The sum rate performance comparison under different power budgets.]{\includegraphics[width=0.325\textwidth]{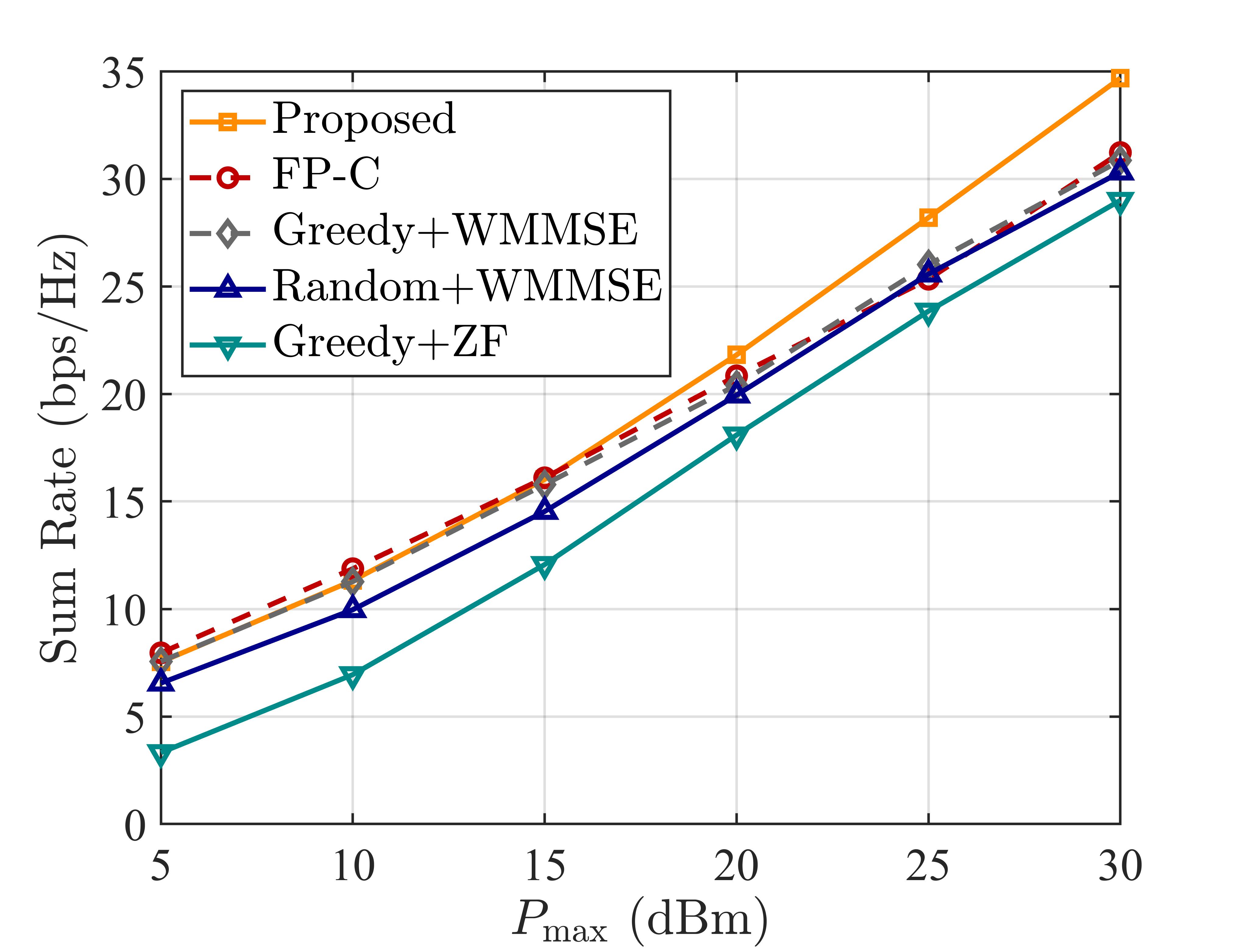}%
\label{MAFig1a}}
\subfigure[The sum rate performance comparison under different $M$'s.]{\includegraphics[width=0.325\textwidth]{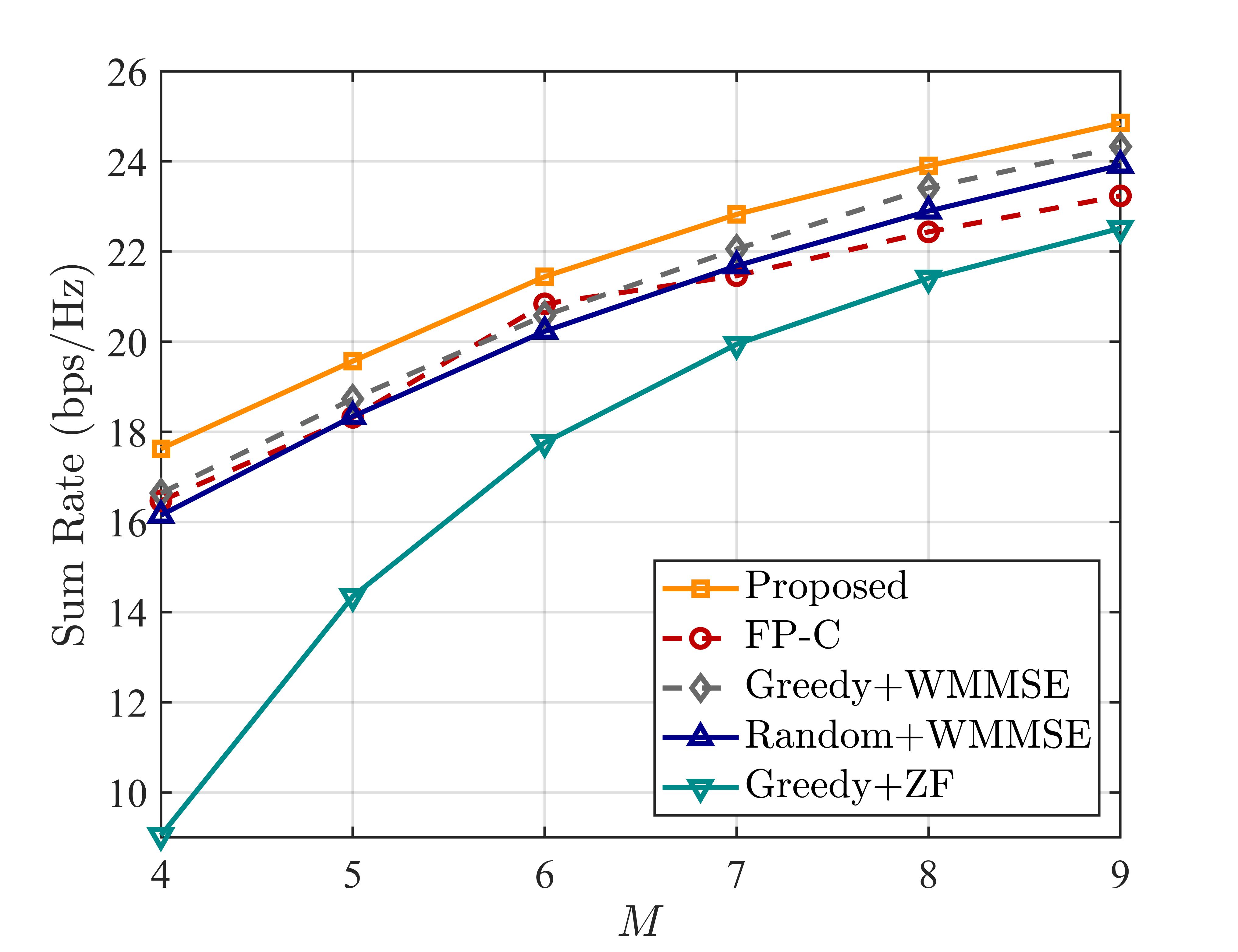}\label{MAFig4}}
\subfigure[The sum rate performance comparison under different $N$'s.]{\includegraphics[width=0.325\textwidth]{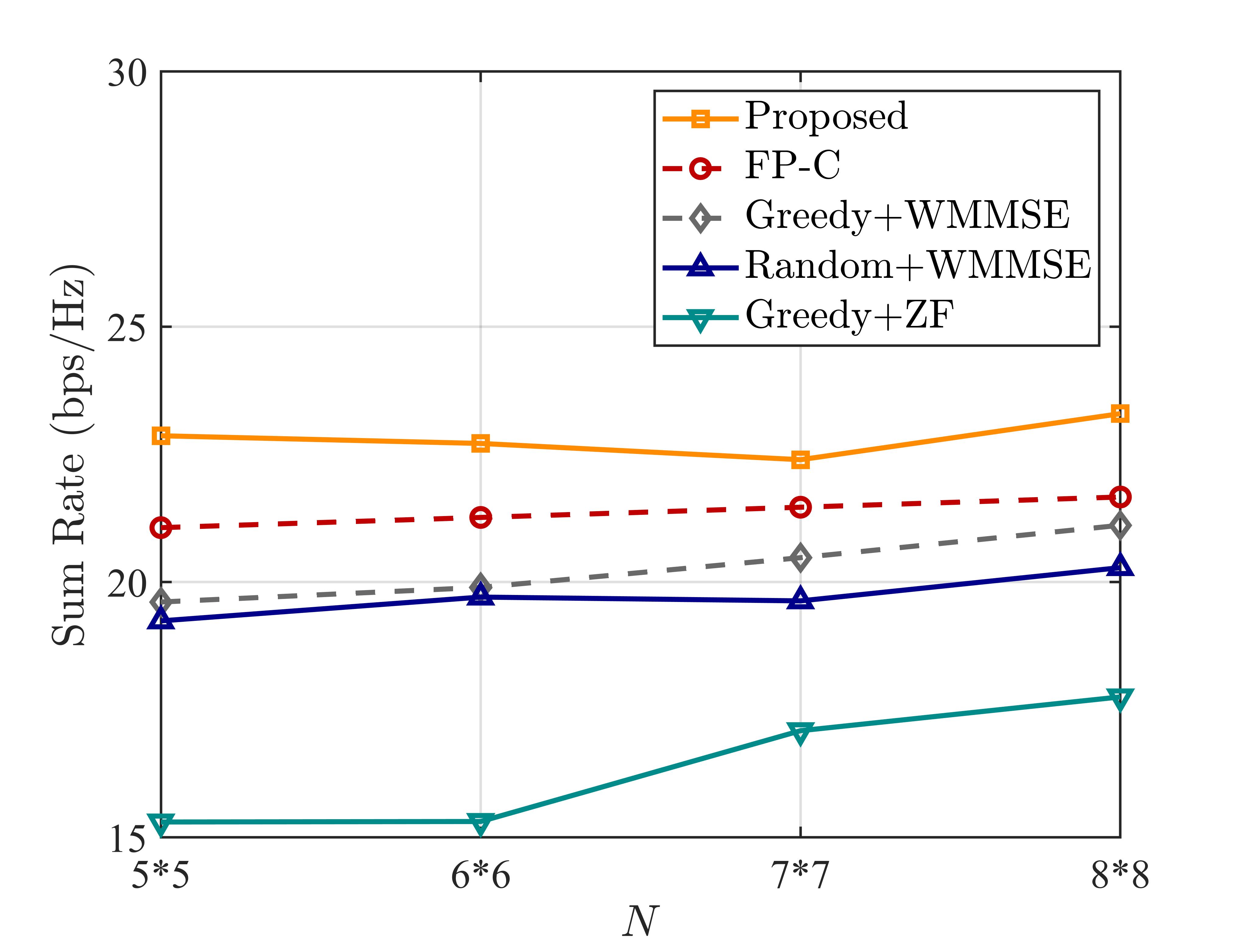}\label{MAFig5}}
\caption{The sum rate performance comparison under different power budgets, $M$'s, and $N$'s.}
\label{MAFig1}
\vspace{-0.15in}
\end{figure*}

\textit{3) Performance Evaluation:} 
We first compare the sum rate performance under different power budgets. In Fig.~\ref{MAFig1}(a), with $M=6$ and $N=49$, it can be observed that \emph{Random+WMMSE} and \emph{Greedy+ZF} perform badly due to their heuristic antenna positioning and beamforming strategies, respectively. Moreover, the proposed DL framework outperforms all baselines particularly at higher transmit power. This demonstrates the great capability of the proposed framework in mitigating interference, particularly when inter-user interference becomes stronger. 

Next, we evaluate the performance of the proposed DL framework under different system settings. We first set $P_{\text{max}}=20$ dBm and $N=49$, and plot the sum rate against $M$ in Fig.~\ref{MAFig4}.  As observed, the proposed DL framework achieves highest sum rates across different values of $M$. Next, in Fig.~\ref{MAFig5}, we plot the sum rate against $N$ when $P_{\text{max}}=20$~dBm and $M=6$. We can see that the proposed DL framework achieves superior performance across different values of $N$. In contrast to heuristic positioning methods such as \emph{Greedy} and \emph{Random} that perform poorly, our proposed method achieves the best performance no matter when the number of CPs is limited, or when the number of CPs is sufficient but with more challenging constraints.

\begin{figure}[t]
\centering
\includegraphics[width=0.3\textwidth]{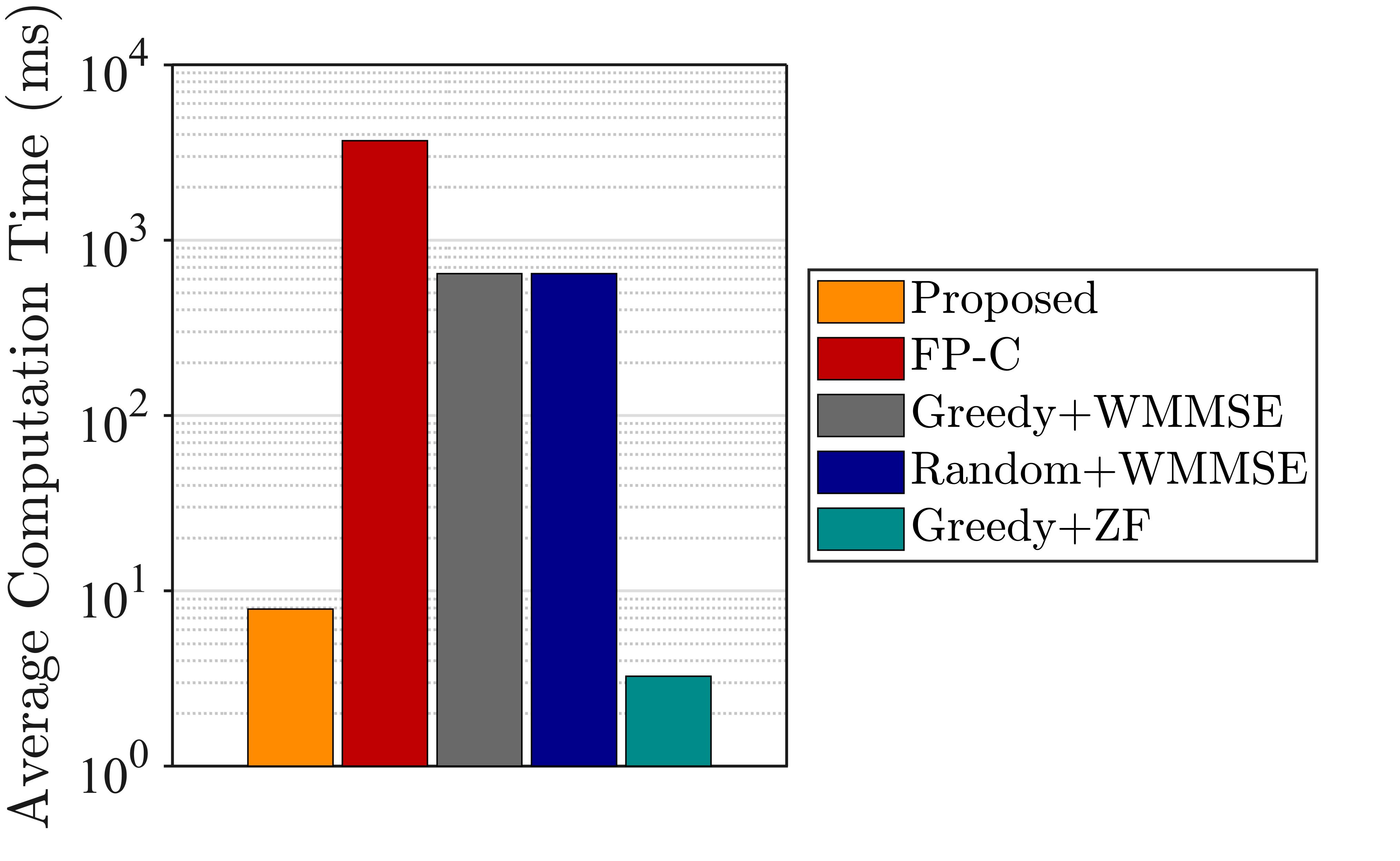}
\caption{The average computation time comparison when $M=6$ and $N=49$.}
\label{MAFig1b}
\end{figure}

We then compare the average computation time of various approaches. It can be observed from Fig.~\ref{MAFig1b} that the computation time of the proposed DL framework is substantially shorter than the iterative optimization-based methods, including \emph{WMMSE} and \emph{FP-C}.

\rev{Finally, to evaluate the optimality of our proposed DL framework, we compare it with an exhaustive search baseline in a small-scale MA-aided system with $N = 25$ and $P_{\max}=20~\text{dBm}$. As shown in Table~\ref{tab:exhaustive_tabel}, the proposed framework achieves over $90\%$ of the sum rate obtained by Exhaustive Search + ZF and consistently outperforms all other benchmarks, confirming that our proposed DL framework can achieve near-optimal performance.}

\begin{table}[t]
  \centering
  \caption{Sum Rate Comparison under different $M$}
  \label{tab:exhaustive_tabel}
  \begin{tabular}{lcccc}
    \toprule
    Method & $M=6$ & $M=7$ & $M=8$ & $M=9$ \\
    \midrule
    Exhaustive Search+ZF & {24.40} & {25.29} & {26.01} & {26.60}  \\
    \rowcolor{pink!30}
    Proposed        & \makecell{22.82 \\ (93.52\%)} & \makecell{22.86 \\ (90.41\%)} & \makecell{23.90 \\ (91.89\%)} & \makecell{24.85 \\ (93.42\%)} \\
    FP-C            & \makecell{20.84 \\ (85.39\%)} & \makecell{21.47 \\ (84.87\%)} & \makecell{22.44 \\ (86.27\%)} & \makecell{23.24 \\ (87.34\%)} \\
    Greedy+WMMSE    & \makecell{19.61 \\ (80.35\%)} & \makecell{20.58 \\ (81.38\%)} & \makecell{22.06 \\ (84.80\%)} & \makecell{23.41 \\ (88.01\%)} \\
    Random+WMMSE    & \makecell{19.24 \\ (78.84\%)} & \makecell{20.23 \\ (79.99\%)} & \makecell{21.68 \\ (83.37\%)} & \makecell{22.90 \\ (86.07\%)} \\
    Greedy+ZF    & \makecell{17.76 \\ (72.78\%)} & \makecell{19.61 \\ (77.55\%)} & \makecell{21.40 \\ (82.29\%)} & \makecell{22.77 \\ (85.58\%)} \\
    \bottomrule
  \end{tabular}
\end{table}

\section{Conclusion}
In this paper, we have proposed a DL framework for solving a general class of mixed-discrete wireless resource allocation problems with discrete constraints. A general problem formulation has been established and a DVLN with probabilistic formulation has been designed to handle discrete variables. It employs an encoder-decoder architecture with a sequential decoding, enabling it to rigorously enforce the coupled discrete constraints while satisfying the non-SPSD property on discrete solutions. With a CVLN developed to predict the solutions of continuous variables, the DVLN and CVLN are jointly trained in an unsupervised manner. Simulation results on two representative case studies have demonstrated the superior performance with significantly less computation time compared to existing model-based and DL-based approaches.

\bibliographystyle{IEEEtran}
\bibliography{ref}

\newpage

\section*{Supplementary Materials}
{\appendices 
\section{Update Mechanism of ENGNN}
\label{APP1}
The ENGNN comprises two types of nodes: transmitter (TX) nodes and receiver (RX) nodes, with cardinalities $N_{\text{T}}$ and $N_{\text{R}}$, respectively. Define the TX-node feature and the RX-node feature in the $l$-th layer as $\mathbf{f}_{\text{TX}, n_{\text{t}}}^{[l]}$ and $\mathbf{f}_{\text{RX}, n_{\text{r}}}^{[l]}$, and there corresponding edge features as $\mathbf{e}_{n_{\text{t}}n_{\text{r}}}^{[l]}$. At the $l$-th layer, the node and edge features are updated by
\begin{equation}
\label{ENGNNUE}
\begin{aligned}
\mathbf{f}_{\text{TX},n_{\text{t}}}^{[l]}= & \text{MLP}_{2}^{[l]} \left(\mathbf{f}_{\text{TX},n_{\text{t}}}^{[l-1]}, \frac{1}{N_{\text{R}}}\sum_{n_{\text{r}}=1}^{N_{\text{R}}}\text{MLP}_{1}^{[l]}\left(\mathbf{f}_{\text{RX},n_{\text{r}}}^{[l-1]}, \mathbf{e}_{n_{\text{t}}n_{\text{r}}}^{[l-1]}\right)\right), \\
&\forall n=1,\dots,N_{\text{T}},
\end{aligned}
\end{equation}
\begin{equation}
\label{ENGNNAP}
\begin{aligned}
\mathbf{f}_{\text{RX},n_{\text{r}}}^{[l]}= & \text{MLP}_{4}^{[l]}\left(\mathbf{f}_{\text{RX},n_{\text{r}}}^{[l-1]}, \frac{1}{N_{\text{T}}}\sum_{n_{\text{t}}=1}^{N_{\text{T}}}\text{MLP}_{3}^{[l_{\text{E}}]}\left(\mathbf{f}_{\text{TX},n_{\text{t}}}^{[l-1]}, \mathbf{e}_{n_{\text{t}}n_{\text{r}}}^{[l-1]}\right)\right), \\
& \forall n_{\text{r}}=1,\dots,N_{\text{R}},
\end{aligned}
\end{equation}
\begin{equation}
\label{ENGNNEdge}
\begin{aligned}
\mathbf{e}_{n_{\text{t}}n_{\text{r}}}^{[l]}=&\text{MLP}_{7}^{[l]} \left(\mathbf{e}_{n_{\text{t}}n_{\text{r}}}^{[l-1]}, \frac{1}{N_{\text{T}}}\sum_{n'=1}^{N_{\text{T}}}\text{MLP}_{5}^{[l]}\left(\mathbf{e}^{[l-1]}_{n'n_{\text{r}}},\mathbf{f}_{\text{RX},n_{\text{r}}}^{[l-1]}\right),\right.\\
&\left.\frac{1}{N_{\text{R}}}\sum_{n'=1}^{N_{\text{R}}}\text{MLP}_{6}^{[l]}\left(\mathbf{e}_{n_{\text{t}}n'}^{[l-1]},\mathbf{f}_{\text{TX},n_{\text{t}}}^{[l-1]}\right)\right),\\
& ~\forall n_{\text{t}}=1,\dots,N_{\text{T}}, ~n_{\text{r}}=1,\dots,N_{\text{R}}.
\end{aligned}
\end{equation}
where $\text{MLP}_{1}^{[l]}$ - $\text{MLP}_{7}^{[l]}$ are 7 MLPs.

\section{Design of $\mathcal{F}_{\text{ctx}}$ for Case Study 1}
\label{APP2}
In the $t$-th decoding step, the edge features of $\mathcal{F}_{\text{ctx}}(\cdot,\cdot,\cdot)$ are initialized as
\begin{equation}
\begin{split}
\bar{\mathbf{e}}_{kl,t}^{[0]}&=\begin{cases}
\text{MLP}_1^{\text{X}}\left(\mathbf{h}_{kl},\mathbf{r}_{kl}\right),~\forall(k,l)\in\mathcal{A}_{t-1}, \\
\text{MLP}_2^{\text{X}}\left(\mathbf{h}_{kl},\mathbf{r}_{kl}\right),~\forall(k,l)\notin\mathcal{A}_{t-1},
\end{cases} \\
\end{split}
\label{Dinitialize}
\end{equation}
where $\text{MLP}_1^{\text{X}}$ and $\text{MLP}_2^{\text{X}}$ are two distinct MLPs, and the node features $\bar{\mathbf{f}}_{\text{UE},k,t}^{[0]}$ and $\bar{\mathbf{f}}_{\text{AP},l,t}^{[0]}$ are initialized as zero. Furthermore, the node and edge features are updated as
\begin{equation}
\label{DUE}
\bar{\mathbf{f}}_{\text{UE},k,t}^{[l_{\text{X}}]}= {\text{MLP}_{3}^{\text{X},[l_{\text{X}}]}}\left(\bar{\mathbf{f}}_{\text{UE},k,t}^{[l_{\text{X}}-1]},\frac{1}{L}\sum_{l=1}^{L}\mathbf{g}_{kl,t}^{[l_{\text{X}}]}\right), ~\forall k\in\mathcal{K},
\end{equation}
\begin{equation}
\label{DAP}
\bar{\mathbf{f}}_{\text{AP},l,t}^{[l_{\text{X}}]}= {\text{MLP}_{4}^{\text{X},[l_{\text{X}}]}}\left(\bar{\mathbf{f}}_{\text{AP},l,t}^{[l_{\text{X}}-1]},\frac{1}{K}\sum_{k=1}^{K}\mathbf{n}_{kl,t}^{[l_{\text{X}}]}\right), ~\forall l\in\mathcal{L},
\end{equation}
\begin{equation}
\label{Ddge}
\begin{split}
\bar{\mathbf{e}}_{kl,t}^{[l_{\text{X}}]} &= \begin{cases}
\text{MLP}_{5}^{\text{X},[l_{\text{X}}]}\left(\bar{\mathbf{e}}_{kl,t}^{[l_{\text{X}}]},\bar{\mathbf{f}}^{[l_{\text{X}}]}_{\text{UE},k,t},\bar{\mathbf{f}}^{[l_{\text{X}}]}_{\text{AP},l,t} \right),~\forall (k,l)\in\mathcal{A}_{t-1}, \\
\text{MLP}_{6}^{\text{X},[l_{\text{X}}]}\left(\bar{\mathbf{e}}_{kl,t}^{[l_{\text{X}}]},\bar{\mathbf{f}}^{[l_{\text{X}}]}_{\text{UE},k,t},\bar{\mathbf{f}}^{[l_{\text{X}}]}_{\text{AP},l,t} \right) ,~\forall (k,l)\notin\mathcal{A}_{t-1},
\end{cases} \\
\end{split}
\end{equation}
where $\text{MLP}^{\text{X},[l_{\text{X}}]}_{3}$-$\text{MLP}^{\text{X},[l_{\text{X}}]}_{6}$ are 4 MLPs, and $\mathbf{g}_{kl,t}^{[l_{\text{X}}]}$ and $\mathbf{n}_{kl,t}^{[l_{\text{X}}]}$ denote the messages aggregated from edges to nodes, which are computed as
\begin{equation}
\begin{split}
\mathbf{g}_{kl,t}^{[l_{\text{X}}]}= \begin{cases}
\text{MLP}_{7}^{\text{X},[l_{\text{X}}]}\left(\bar{\mathbf{f}}^{[l_{\text{X}}-1]}_{\text{AP},l,t},\bar{\mathbf{e}}_{kl,t}^{[l_{\text{X}}-1]} \right),\forall (k,l)\in\mathcal{A}_{t-1}, \\
\text{MLP}_{8}^{\text{X},[l_{\text{X}}]}\left(\bar{\mathbf{f}}^{[l_{\text{X}}-1]}_{\text{AP},l,t},\bar{\mathbf{e}}_{kl,t}^{[l_{\text{X}}-1]} \right),\forall (k,l)\notin\mathcal{A}_{t-1},
\end{cases} \\
\end{split}
\label{nAP}
\end{equation}
\begin{equation}
\begin{split}
\mathbf{n}_{kl,t}^{[l_{\text{X}}]}= \begin{cases}
\text{MLP}_{9}^{\text{X},[l_{\text{X}}]}\left(\bar{\mathbf{f}}^{[l_{\text{X}}-1]}_{\text{UE},k,t},\bar{\mathbf{e}}_{kl,t}^{[l_{\text{X}}-1]} \right),~\forall (k,l)\in\mathcal{A}_{t-1}, \\
\text{MLP}_{10}^{\text{X},[l_{\text{X}}]}\left(\bar{\mathbf{f}}^{[l_{\text{X}}-1]}_{\text{UE},k,t},\bar{\mathbf{e}}_{kl,t}^{[l_{\text{X}}-1]} \right),~\forall (k,l)\notin\mathcal{A}_{t-1},
\end{cases} \\
\end{split}
\label{nUE}
\end{equation}
where $\text{MLP}^{\text{X},[l_{\text{X}}]}_{7}$-$\text{MLP}^{\text{X},[l_{\text{X}}]}_{10}$ are 4 additional MLPs. The associated and candidate edges are modeled as distinct edge types, each processed by dedicated MLPs to capture their unique characteristics. Finally, after $L_{\text{X}}$ layers, the graph-level context vector is obtained by aggregating node and edge representations:
\begin{equation}
\begin{aligned}\mathbf{c}^{\star}_{t}=&\frac{1}{K}\sum_{k=1}^{K}\text{FC}_{1}^{\text{X}}({\bar{\mathbf{f}}}_{\text{UE},k,t}^{[L_\text{X}]})+\frac{1}{L}\sum_{l=1}^{L}\text{FC}_{2}^{\text{X}}({\bar{\mathbf{f}}}_{\text{AP},l,t}^{[L_\text{X}]}) \\
&+\frac{1}{KL}\sum_{k=1}^{K}\sum_{l=1}^{L}\text{FC}_{3}^{\text{X}}({\bar{\mathbf{e}}}_{kl,t}^{[L_\text{X}]}),
\end{aligned}
\end{equation} 
where $\text{FC}_1^{\text{X}}(\cdot)-\text{FC}_3^{\text{X}}(\cdot)$ are single-layer fully connected networks.

\section{Design of the critic-network for Case Study 1}
In training period, the critic-network $\hat{U}(\cdot)$ is introduced to assist the training of $\mathcal{F}_\text{A}(\cdot)$, which takes the channel $\mathbf{h}$ as the input, and outputs the estimated sum rate. Its architecture closely mirrors that of $\mathcal{G}_{\text{E}}(\cdot)$, with the only difference in the output layer. The output layer of $\hat{U}(\cdot)$ is expressed as
\begin{equation}
\begin{aligned}
\hat{U}(\mathbf{h})=&\text{ReLU}\left[\frac{1}{K}\sum_{k=1}^{K}\text{FC}_{1}^{\text{C}}({\tilde{\mathbf{f}}}_{\text{UE},k}^{[L_\text{C}]})+\frac{1}{L}\sum_{l=1}^{L}\text{FC}_{2}^{\text{C}}({\tilde{\mathbf{f}}}_{\text{AP},l}^{[L_\text{C}]}) \right.\\
&\left.+\frac{1}{KL}\sum_{k=1}^{K}\sum_{l=1}^{L}\text{FC}_{3}^{\text{C}}({\tilde{\mathbf{e}}}_{kl}^{[L_\text{C}]})\right],
\end{aligned}
\label{CriticCF}
\end{equation} 
where $\text{FC}_{1}^{\text{C}}(\cdot)$-$\text{FC}_{3}^{\text{C}}(\cdot)$ are single-layer fully connected networks, $L_{\text{C}}$ denotes the number of hidden layers of the critic-network, $\tilde{\mathbf{f}}_{\text{UE},k}^{[L_{\text{C}}]}$, $\tilde{\mathbf{f}}_{\text{AP},l}^{[L_{\text{C}}]}$, and $\tilde{\mathbf{e}}_{kl}^{[L_{\text{C}}]}$ correspond to the features of the $k$-th UE-node, the $l$-th AP-node, and their associated edge features at the final hidden layer, respectively, and the ReLU activation is applied to ensure the non-negative output.

\section{Design of $\mathcal{F}_{\text{w}}(\cdot,\cdot)$ for Case Study 2}
In this subsection, we introduce the design of $\mathcal{F}_{\text{w}}(\cdot,\cdot)$. First, we note the optimal solution of the beamformer in this case study~\cite{OptimMUMISO}
\begin{equation}
\label{MAbeamstruc}
\begin{aligned}
{\mathbf{w}}_{k}=&\sqrt{p_{k}} \frac{\left(\mathbf{I}_{M}+\sum_{i=1}^{K} \frac{\mu_{i}}{\sigma_{k}^{2}} \mathbf{h}_{i}(\mathbf{b}) \mathbf{h}_{i}(\mathbf{b}) ^{H}\right)^{-1} \mathbf{h}_{k}(\mathbf{b})}{\left\|\left(\mathbf{I}_{M}+\sum_{i=1}^{K} \frac{\mu_{i}}{\sigma_{k}^{2}} \mathbf{h}_{i}(\mathbf{b}) \mathbf{h}_{i}(\mathbf{b}) ^{H}\right)^{-1} \mathbf{h}_{k}(\mathbf{b})\right\|}, \\
&~\forall k \in \mathcal{K},
\end{aligned}
\end{equation}
where $\boldsymbol{\mu}\triangleq[\mu_1,\mu_2,\dots,\mu_{K}]^T\in\mathbb{R}_{+}^{K}$ and $\boldsymbol{p}\triangleq[p_1,p_2,\dots,p_{K}]^T\in\mathbb{R}_{+}^{K}$ are unknown parameters and should satisfy $\sum_{k=1}^{K}\mu_{k}=\sum_{k=1}^{K}p_{k}=P_{\text{max}}$ such that the maximum power constraints can be satisfied. Exploiting this solution structure can simplify the mapping to be learned by the NN, by decreasing the number of unknown parameters from $2KM$ to $2K$. 

The remaining task is to predict these low-dimensional parameters. For this purpose, we adopt the ENGNN, which consists of two types of nodes, i.e., $M$ MA-nodes and $K$ UE-nodes, and there exists an edge between each MA-node and each UE-node. The edge features are initialized as
\begin{equation}
\hat{\mathbf{e}}^{[0]}_{km}=\text{MLP}^{\text{W}}_{1}([\mathbf{h}_{k}(\mathbf{b})]_m),~\forall k \in\mathcal{K},~m\in \mathcal{M},
\end{equation}
and the node features $\hat{\mathbf{f}}^{[0]}_{\text{MA},m}$ and $\hat{\mathbf{f}}^{[0]}_{\text{UE},k}$ are initialized as zero. After $L_{\text{W}}$ layers node and edge updating, we obtain $\boldsymbol{\mu}$ and $\boldsymbol{p}$ by
\begin{subequations}
\begin{align}
&[{\mu}_k,{p}_k]^T= \text{FC}^{\text{W}}(\hat{\mathbf{f}}^{[L_{\text{W}}]}_{\text{UE},k}),~~\forall k \in \mathcal{K}, \\
&\boldsymbol{\mu} \gets P_{\max}\cdot\text{softmax}({\boldsymbol{\mu}}), \\
&\boldsymbol{p} \gets P_{\max}\cdot\text{softmax}({\boldsymbol{p}}), 
\end{align}
\end{subequations}
where $\text{FC}^{\text{W}}(\cdot)$ is a single-layer fully-connected network, and the softmax activation is employed to normalize $\boldsymbol{\mu}$ and $\boldsymbol{p}$. Finally, $\boldsymbol{\mu}$ and $\boldsymbol{p}$ are substituted to (\ref{MAbeamstruc}) to get $\mathbf{w}$.

\rev{\section{Generalization Performance}
We evaluate the generalization performance of the proposed DL framework. Specifically, in Case Study 1, we train our NN model under $K=20$ users, $L=8$ APs, and subsequently tested it under varying $K$ and $L$. Moreover, in Case Study 2, we train the model with $M=6$ MAs and $N=49$ CPs, and test it across different $M$ and $N$. As observed in Fig.~\ref{fig:CFGen} and Fig.~\ref{fig:MAGen}, our framework achieves stable sum rate performance despite the mismatched system dimensions, thereby validating its robust generalization ability.}

\begin{figure}[t]
\centering
\subfigure[Generalization to $K$, with $L=8$]{\includegraphics[width=0.4\textwidth]{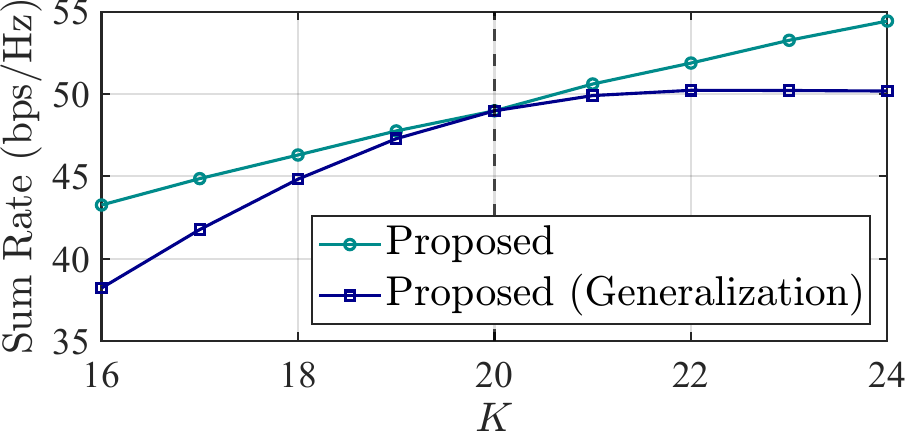}%
\label{fig:CFGen_K}}
\hfil 
\subfigure[Generalization to $L$, with $K=20$]{\includegraphics[width=0.4\textwidth]{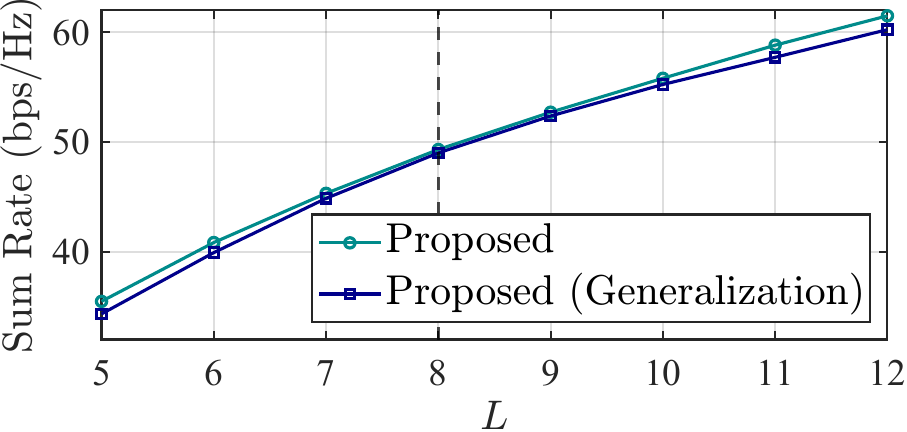}%
\label{fig:CFGen_L}}
\caption{Generalization to $K$ and $L$.}
\label{fig:CFGen}
\end{figure}

\begin{figure}[t]
\centering
\subfigure[Generalization to $M$,with $N=49$]{\includegraphics[width=0.4\textwidth]{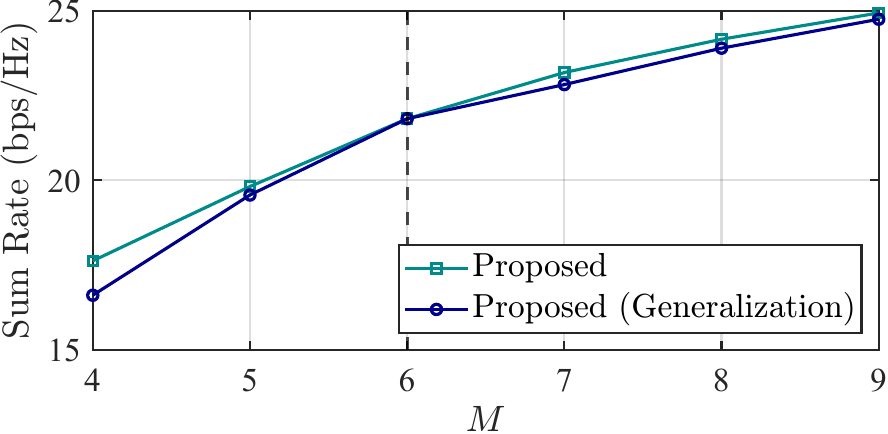}\label{fig:MAGen_M}}
\subfigure[Generalization to $N$, with $M=6$]{\includegraphics[width=0.4\textwidth]{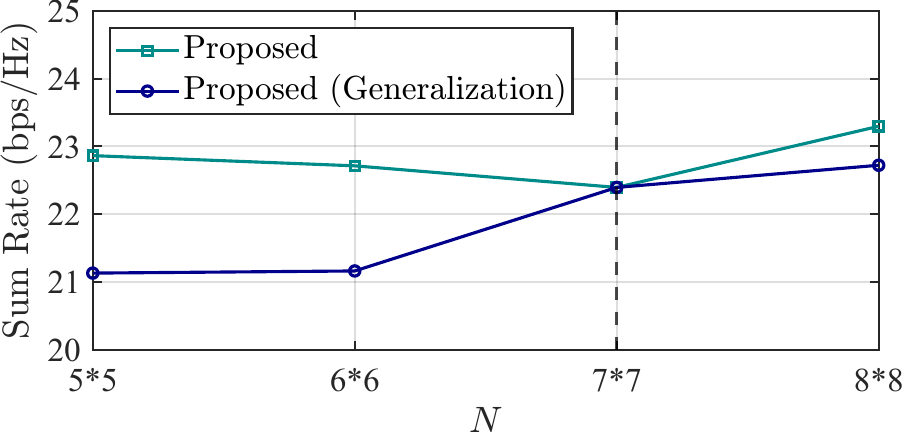}\label{fig:MAGen_N}}
\caption{Generalization to $M$ and $N$.}
\label{fig:MAGen}
\end{figure}
}
\end{document}